\documentclass[lettersize,journal]{IEEEtran}

\usepackage{graphicx}
\usepackage{amsmath}
\usepackage{amssymb}
\usepackage{amsthm}
\usepackage{bbm}
\usepackage{subfigure}
\usepackage{times}
\usepackage{latexsym}
\usepackage{microtype}
\usepackage{color}
\usepackage{float}
\usepackage{tabularx,multirow}
\usepackage{hyperref}
\usepackage{breakurl}
\makeatletter
\def\UrlAlphabet{%
      \do\a\do\b\do\c\do\d\do\e\do\f\do\g\do\h\do\i\do\j%
      \do\k\do\l\do\m\do\n\do\o\do\p\do\q\do\r\do\s\do\t%
      \do\u\do\v\do\w\do\x\do\y\do\z\do\A\do\B\do\C\do\D%
      \do\E\do\F\do\G\do\H\do\I\do\J\do\K\do\L\do\M\do\N%
      \do\O\do\P\do\Q\do\R\do\S\do\T\do\U\do\V\do\W\do\X%
      \do\Y\do\Z}
\def\UrlDigits{\do\1\do\2\do\3\do\4\do\5\do\6\do\7\do\8\do\9\do\0}
\g@addto@macro{\UrlBreaks}{\UrlOrds}
\g@addto@macro{\UrlBreaks}{\UrlAlphabet}
\g@addto@macro{\UrlBreaks}{\UrlDigits}

\usepackage{makecell}
\usepackage{booktabs}

\usepackage{bm}

\usepackage{ulem}

\usepackage{multirow}
\usepackage{multicol}
\usepackage{array}
\usepackage[ruled,linesnumbered]{algorithm2e}

\usepackage{amsmath}

\ifCLASSOPTIONcompsoc
  \usepackage[nocompress]{cite}
\else
  \usepackage{cite}
\fi
\ifCLASSINFOpdf
\else
\fi
\begin{document}


\title{Spatio-Temporal Fuzzy-oriented Multi-Modal Meta-Learning for Fine-grained Emotion Recognition}

\author{Jingyao~Wang,
        Wenwen~Qiang,
        Changwen~Zheng,
        and Fuchun~Sun,~\IEEEmembership{Fellow,~IEEE}
\IEEEcompsocitemizethanks{\IEEEcompsocthanksitem J. Wang, W. Qiang, and C. Zheng are with University of Chinese Academy of Sciences, Beijing, China. They are also with National Key Laboratory of Space Integrated Information System, Institute of Software Chinese Academy of Sciences, Beijing, China. E-mail: {wangjingyao2023, qiangwenwen, changwen}@iscas.ac.cn.
\IEEEcompsocthanksitem F. Sun is with Department of Computer Science and Technology, Tsinghua University, Beijing, China. E-mail: fcsun@tsinghua.edu.cn.
}}

\markboth{IEEE Transactions on Systems, Man and Cybernetics: Systems}{Wang \MakeLowercase{\textit{et al.}}: Spatio-Temporal Fuzzy-oriented Multi-Modal Meta-Learning for Fine-grained Emotion Recognition}


\IEEEtitleabstractindextext{%
\begin{abstract}
Fine-grained emotion recognition (FER) plays a vital role in various fields, such as disease diagnosis, personalized recommendations, and multimedia mining. However, existing FER methods face three key challenges in real-world applications: (i) they rely on large amounts of continuously annotated data to ensure accuracy since emotions are complex and ambiguous in reality, which is costly and time-consuming; (ii) they cannot capture the temporal heterogeneity caused by changing emotion patterns, because they usually assume that the temporal correlation within sampling periods is the same; (iii) they do not consider the spatial heterogeneity of different FER scenarios, that is, the distribution of emotion information in different data may have bias or interference. To address these challenges, we propose a Spatio-Temporal Fuzzy-oriented Multi-modal Meta-learning framework (ST-F2M). Specifically, ST-F2M first divides the multi-modal videos into multiple views, and each view corresponds to one modality of one emotion. Multiple randomly selected views for the same emotion form a meta-training task. Next, ST-F2M uses an integrated module with spatial and temporal convolutions to encode the data of each task, reflecting the spatial and temporal heterogeneity. Then it adds fuzzy semantic information to each task based on generalized fuzzy rules, which helps handle the complexity and ambiguity of emotions. Finally, ST-F2M learns emotion-related general meta-knowledge through meta-recurrent neural networks to achieve fast and robust fine-grained emotion recognition. Extensive experiments show that ST-F2M outperforms various state-of-the-art methods in terms of accuracy and model efficiency. In addition, we construct ablation studies and further analysis to explore why ST-F2M performs well. 
\end{abstract}

}

\maketitle

\IEEEdisplaynontitleabstractindextext

%
\IEEEpeerreviewmaketitle


\emph{This work has been submitted to the IEEE for possible publication. Copyright may be transferred without notice, after which this version may no longer be accessible.}

\section{Introduction}
\label{sec:1}
\IEEEPARstart{F}{ine-grained} emotion recognition (FER) uses physiological signals to model the changes in emotions and provide refined descriptions. FER has various applications in multiple fields such as disease diagnosis \cite{chen2021multimodal,li2023graphcfc}, social media analysis \cite{kabir2021emocov,zhang2021recognition}, and market research \cite{zhang2021customer}. Take the application of FER in disease diagnosis as an example, it can assist in diagnosing depression, schizophrenia, and other psychological diseases, and also adjust the medical plan and service based on the patient's emotional feedback \cite{huynh2020self, li2022intelligent,chen2021multimodal, liu2021subtype}.

\begin{figure}
     \centering
     \subfigure[Data dependency]{\includegraphics[width=0.240\textwidth]{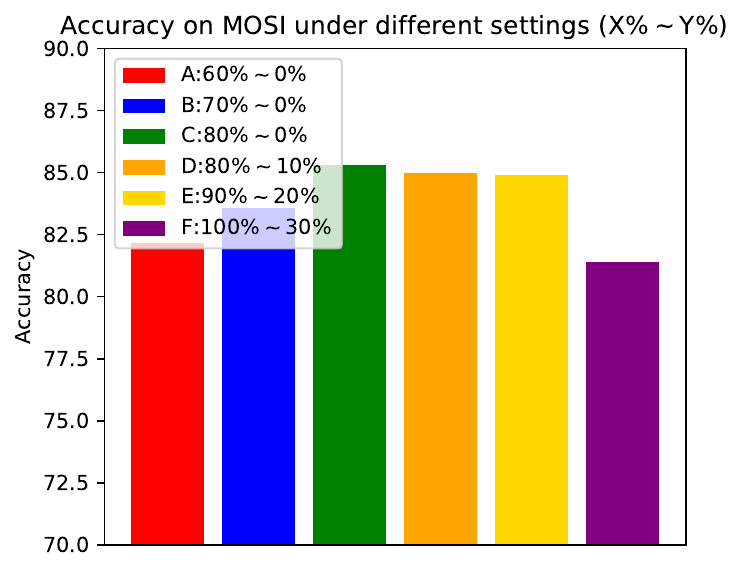}\label{fig:intro_1}}
     \subfigure[Temporal heterogeneity]{\includegraphics[width=0.240\textwidth]{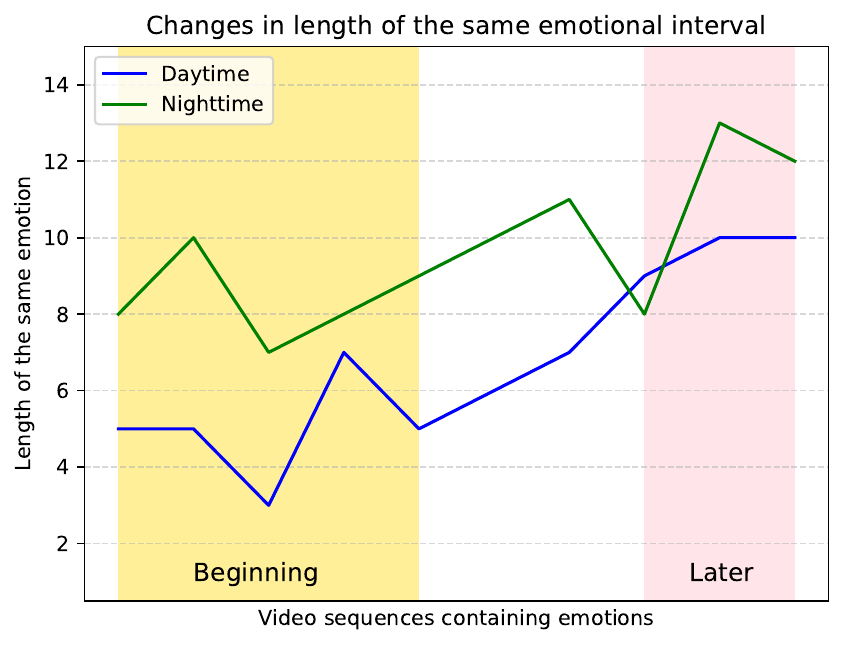}\label{fig:intro_2}}
     \subfigure[Spatial heterogeneity]{\includegraphics[width=0.5\textwidth]{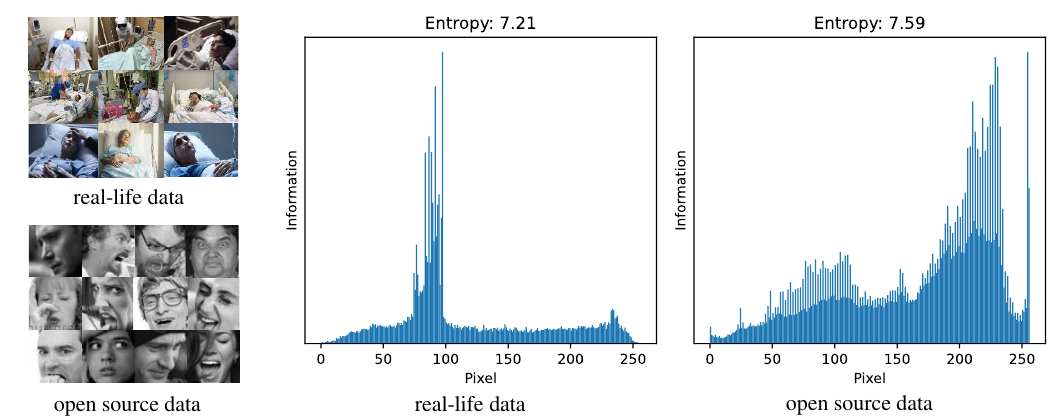}\label{fig:intro_3}}
    \caption{Illustration of our motivation, i.e., the three key challenges in fine-grained emotion recognition.}
    \label{fig:1}
\end{figure}

Recently, many FER methods have been proposed to simulate the dynamics of human emotional states \cite{wei2021fine,debnath2023disturbance,li2024brain}. Unlike discrete emotion recognition (DER) \cite{zhao2021combining,ji2024mlg} which only assigns a single discrete emotion label to an image, FER can capture the nature of human emotions changing over time by using fine-grained segments (usually 0.5 seconds to 4 seconds) \cite{xiao2022user,zhang2022few}. Therefore, FER can achieve higher prediction accuracy than DER \cite{tolosa2021challenges}. Although significant efforts have been made to improve FER performance, existing methods still face three key challenges, especially in real-life long videos.

\textbf{The first challenge} is relying on a large amount of continuous annotated data to ensure accuracy due to the complexity and ambiguity of emotions \cite{wei2021fine,zhu2022multimodal}. We analyze the performance of UniMSE \cite{hu2022unimse} under different data sizes and sample defects. Figure \ref{fig:intro_1} shows the results of six data groups conducted from MOSI \cite{zadeh2016mosi}, denoted as $A \to F (X\% \sim Y\%)$. For each group, $X\%$ and $Y\%$ denote the data size of the corresponding group and the area of defects in each sample, respectively. The results show that (i) the model performs better in group C than A and B, indicating that more data leads to a performance improvement; and (ii) the model performs the worst in group F, illustrating that inaccurate annotations result in larger training errors even with the largest training data. However, in practice, it is challenging to provide a substantial amount of continuous reliable annotation data, which makes experiments costly and time-consuming. Although recent methods \cite{wang2020learning, zhang2022few} have achieved fast convergence on limited data, they still depend on manually crafted features, limiting the model's generalization capability.

\textbf{The second challenge} lies in the use of shared parameter space to model temporal dynamics throughout all time intervals, which makes it hard to accurately capture the temporal heterogeneity within the latent embedding space. In real-world scenarios, the temporal intervals of emotional state changes for different targets can vary over time, e.g., when expressing the same emotion during a conversation, the length of the emotional change interval can differ due to variations in the target's physical exhaustion \cite{poria2019emotion}. This results in temporal heterogeneity, as illustrated in Figure \ref{fig:intro_2}. Some prior work \cite{liu2023feda, liang2020fine} assumes that temporal heterogeneity across the entire time span is static, which is not always held, e.g., there might be a significant difference in the frequency of emotional changes between daytime and nighttime hours shown in Figure \ref{fig:intro_2}.

\textbf{The third challenge} concerns the lack of modeling spatial heterogeneity across different scenarios. Taking Figure \ref{fig:intro_3} as an example, it represents two distinct emotion recognition data sources for disease diagnosis, consisting of the data from real life and the open source data \cite{barsoum2016training}, respectively. It is evident that real data may contain significant occlusions and blurriness, which are sources of interference that open-source datasets overlook. From Figure \ref{fig:1}(c), we can observe substantial differences in the distribution of emotion-related information between these sources. However, most existing models tend to overlook this spatial heterogeneity by employing fixed structures that directly encode topological information and propagate it \cite{xie2019adaptive, liu2023feda}. This approach may not be the most suitable choice for fine-grained emotion recognition tasks due to its inability to adapt to varying spatial characteristics.

To address the above issues, we propose a novel Spatio-Temporal Fuzzy-oriented Multi-modal Meta-learning framework (ST-F2M) for robust fine-grained emotion recognition. To encode spatio-temporal emotional states, we introduce an integrated module with both temporal and spatial convolutions to capture spatio-temporal heterogeneity. To handle complex emotion changes, we adopt generalized fuzzy rules that introduce emotional intensity information into multi-modal data to cope with the complexity and ambiguity of emotions. To enhance model efficiency, our ST-F2M is built upon a meta-recurrent neural network. This network extracts emotion-related information from limited data and converges rapidly. 

In summary, our contributions are as follows:
\begin{itemize}
    \item To the best of our knowledge, we are the first to consider spatio-temporal heterogeneity in real-world fine-grained emotion recognition. It may serve as an inspiration for other practical emotion recognition applications.
    \item We propose the spatio-temporal fuzzy-oriented multi-modal meta-learning framework to achieve robust fine-grained emotion recognition. It addresses three key challenges of real-world FER.
    \item We conduct two fuzzy inference systems based on generalized fuzzy rules which can help us provide emotional intensity information to build FER benchmarks.
    \item Extensive experiments on five benchmark data sets demonstrate the robustness and computational efficiency of our ST-F2M, which outperforms SOTA baselines.
\end{itemize}

\section{Related work}
\label{sec:2}

\subsection{Fine-grained Emotion Recognition}
\label{sec:2.1}

Fine-grained emotion recognition is a key task with many practical applications. It has evolved from physics-based models \cite{albrecht2005mixed,chen2023convolutional} to data-driven models \cite{poria2017review,wang2025design}, and from coarse-grained \cite{poria2019emotion,nie2023long} to fine-grained analysis \cite{liu2023feda}. Different types of data can be used to detect fine-grained emotions, such as body movements \cite{keltner2019emotional}, blood pressure \cite{zhang2016multimodal}, textual information \cite{al2018survey} and facial expressions \cite{ge2022facial}. Among them, facial expressions and textual information are the most convenient and important sources for emotion recognition \cite{poria2017review}, as they are easily captured and abundant in social media.

Traditional methods mainly use handcrafted features which can be divided into two categories, i.e., appearance-based \cite{happy2012real, siddiqi2015human, ghimire2017facial} and geometric-based \cite{ko2018brief, tang2018geometric, liu2020ga}. They contain emotional information, but can only reflect shallow features and are not suitable for structured images. In contrast, deep learning-based methods \cite{krithika2021graph, lo2020mer} can extract more advanced features, and have been a breakthrough in recent decades. Among the most widely used for static emotion recognition are CNNs \cite{lo2020mer, krithika2021graph, liu2023feda}. These approaches can perform optimal graph reasoning, while providing long-range context information. Considering the dynamic change of emotions, some researchers have attempted to exploit recurrent neural networks (RNNs) in emotion recognition \cite{zhang2017facial, zhang2022few, li2023fg}. However, all these methods still face the three challenges as mentioned in Section \ref{sec:1}. Besides, existing methods may have large errors in the process of information transfer when the scenarios are challenging and complex, as shown in Figure \ref{fig:1}(c). In contrast, our ST-F2M utilizes multi-modal meta-learning architecture-driven optimization to eliminate these challenges, achieving robust fine-grained emotion recognition.

\subsection{Generalized Fuzzy Emotion}
\label{sec:2.2}

Human emotions are divided into six basic categories, including happiness, anger, sadness, disgust, fear, and surprise \cite{ekman1992there}. This is used as the foundation for emotion recognition \cite{suhaimi2020eeg, scarpina2022gut, naji2022emo}. However, human emotions often do not exist in isolation, but influence and are intertwined with each other. Existing studies ignore the cross-combination of emotions that are widely present in real life.

The application of generalized fuzzy mathematics provides solutions to the above problem. It can better handle the complexity and diversity of emotions by introducing fuzzy sets and fuzzy rules. Priya et al. \cite{vishnu2019emotion} present a novel geometrical fuzzy-based approach to accurately recognize emotions using fuzzy membership functions. Sreenivas et al. \cite{sreenivas2020group} propose a recurrent fuzzy neural network (RFNN) for emotion classification. Liliana et al. \cite{liliana2016human} display the degree of fuzzy cluster on eight face emotions based on semi-supervised Fuzzy C-means (FCM). Although these works proposed a definition of generalized fuzzy emotions, they only clustered emotions of different intensities without considering the diversity of emotions. In this paper, we aim to utilize fuzzy math and conduct generalized fuzzy rules to help quickly identify mixed emotions.

\subsection{Meta-Learning for Emotion Recognition}
\label{sec:2.3}

Humans can learn from a few samples and easily develop an understanding of new tasks, thus interacting with the external world. For example, humans can quickly identify different types of apples in real life after only seeing a picture of an apple. Meta-learning aims at equipping machines with this ability to adapt to new scenarios effectively \cite{mahadevkar2022review}. Meta-learning methods can be categorized into three types: optimization-based methods \cite{finn2017model, nichol2018reptile, raghu2019rapid}, model-based methods \cite{santoro2016meta, obamuyide2019meta}, and metric-based methods \cite{snell2017prototypical, sung2018learning, chen2020variational}. They all rely on shared structures to extract meta-knowledge, enabling excellent performance on new tasks, and are widely applied in few-shot learning.

Recent studies have attempted to employ meta-learning for emotion recognition \cite{poria2017review}. Tang et al. \cite{tang2022deep} proposed a meta-transfer learning model aimed at extracting emotional information from electroencephalogram (EEG) data in different environments. Kuruvayil et al. \cite{kuruvayil2022emotion} leveraged the rapid adaptation capability of prototype networks for emotion recognition while mitigating the severe overfitting issue caused by limited samples. Nguyen et al. \cite{nguyen2023meta} introduced an optimization-based PathNet to facilitate the transfer of emotional knowledge from visual to audio domains. Although these works have all employed meta-learning methods to address the challenge of limited training data, they have overlooked the unique challenges posed by real-world data \cite{wei2021fine,barsoum2016training}, such as sample defects, complex environments, and subtle differences in fine-grained emotions. In this study, we are the first to consider and explore the spatio-temporal heterogeneity of fine-grained emotions in the real world, while utilizing meta-learning to unearth the intrinsic factors influencing recognition.

\section{Problem Definition}
\label{sec:3}

In this section, we first introduce the problem settings and notations of meta-learning for FER. 

With a given task distribution $p(\mathcal{T})$, both the meta-training set $\mathcal{D}_{tr}$ and the meta-testing set $\mathcal{D}_{te}$ are randomly drawn from $p(\mathcal{T})$ without any class-level overlap. Within a single training batch, we denote the $N_{tr}$ training tasks as $\left \{ \tau_i \right \}_{i=1}^{N_{tr}} \in \mathcal{D}_{tr}$. Each task $\tau_{i}$ comprises a support set $\mathcal{D}_i^s=(X_i^s,Y_i^s)=  \{ (x_{i,j}^s,y_{i,j}^s)  \}_{j=1}^{N_i^s}$ and a query set $\mathcal{D}_i^q=(X_i^q,Y_i^q)=  \{ (x_{i,j}^q,y_{i,j}^q)  \}_{j=1}^{N_i^q}$. Here, $(x_{i,j}^{\cdot},y_{i,j}^{\cdot})$ represents an individual sample and its corresponding label respectively, while $N_i^{\cdot}$ indicates the number of samples in each set.

Meta-learning can be conceptualized as a two-level optimization problem. In the first level (inner loop), we learn the task-specific model $f^i_{\theta}$ for task $\tau_{i}$ with the support set $\mathcal{D}^{s}{i}$ utilizing the meta-learning model $f_{\theta}$. The objective is:
\begin{equation}
\label{eq:inner}
\begin{array}{l}
    f^i_{\theta} \gets f_{\theta} -\alpha \nabla_{f_{\theta}}\mathcal{L}(X_i^s,Y_i^s,f_{\theta})
\end{array}
\end{equation}
where $\alpha$ is the learning rate. In the second level (outer loop), we update the meta-learning model $f_{\theta}$ by minimizing the average loss computed across multiple tasks, incorporating the query sets $\mathcal{D}^{q}$ and leveraging the knowledge gained from the previously learned task-specific models. The objective is:
\begin{equation}
\label{eq:outer}
\begin{array}{l}
    f_{\theta} \gets f_{\theta}-\beta \nabla_{f_{\theta}}\frac{1}{N_{tr}}\sum_{i=1}^{N_{tr}}\mathcal{L}(X_i^q,Y_i^q,f_\theta ^i)
\end{array}
\end{equation}
where $\beta$ is the learning rate. Note that $f_\theta^i$ is derived by computing the derivative of $f_\theta$, making $\nabla_{f_{\theta}}\frac{1}{N_{tr}}\sum_{i=1}^{N_{tr}}\mathcal{L}(X_i^q,Y_i^q,f_\theta ^i)$ be interpreted as the second-order derivative of $f_\theta$. 

In this study, we will mine meta-knowledge in fine-grained emotion recognition based on this paradigm to improve generalization and transferability.

\begin{table*}
    \begin{minipage}{0.47\linewidth}
        \centering
        \caption{Description of Facial Components. The $fc_i$ in the ``Component'' refers to the $i$-th facial component, e.g., $fc_1$ denotes the left eyebrow, and the ``Attributes'' represents the corresponding attributes and their values.}
        \label{tab:1}
        \resizebox{\linewidth}{!}{
        \begin{tabular}{lll}
        \toprule
        \textbf{Component} & \textbf{Attributes} \\
        \midrule
        $fc_{1}$: left eyebrow & Low-Normal-Raised, (-1)-0-1\\
        $fc_{2}$: right eyebrow  & Low-Normal-Raised, (-1)-0-1\\
        $fc_{3}$: brow crest & Frowning-Normal, 1-0\\
        $fc_{4}$: left eye & Narrow-Normal-Wide, (-1)-0-1\\
        $fc_{5}$: right eye & Narrow-Normal-Wide, (-1)-0-1\\
        $fc_{6}$: nose & Wrinkled-Normal, 1-0\\
        $fc_{7}$: nostril & Dilated-Normal, 1-0\\
        $fc_{8}$: mouth & Normal-Open, 0-1\\
        $fc_{9}$: upper lip & Thin-Normal-Thick, (-1)-0-1\\
        $fc_{10}$: lower lip & Thin-Normal-Thick, (-1)-0-1\\
        $fc_{11}$: left mouth corner & Low-Normal-Raised, (-1)-0-1\\
        $fc_{12}$: right mouth corner & Low-Normal-Raised, (-1)-0-1\\
        \bottomrule
        \end{tabular}}
    \end{minipage}
    \hfill
    \begin{minipage}{0.47\linewidth}
            \centering
            \caption{Examples of FKIS Inference. ``Emotions'' represents the emotion category and corresponding intensity, and ``Coding of $fc_i$'' is the attribute value sequence of the facial components corresponding to the emotion intensity.}
            \label{tab:2}
            \resizebox{0.84\linewidth}{!}{
            \begin{tabular}{ll}
            \toprule
            \textbf{Emotions} & \textbf{Coding of $fc_i$} \\
            \midrule
            Angry-High & 0010010(-1)(-1)(-1)(-1)(-1)\\
            Angry-Medium & (-1)0(-1)000101011\\
            Angry-Low & 0001101(-1)0000\\
            Happy-High & 110110100111\\
            Happy-Medium & 011011000(-1)00\\
            Happy-Low & (-1)010100010(-1)(-1)\\    
            Disgust-High & 11(-1)000001(-1)00\\
            Disgust-Medium & 110000000(-1)00\\
            Fear-Low & 000(-1)(-1)0000011\\
            Sad-High & (-1)(-1)0(-1)(-1)100(-1)(-1)00\\
            Sad-Medium & (-1)(-1)0001000(-1)00\\
            Surprise-Low & 000110000011\\
            \bottomrule
            \end{tabular}}
    \end{minipage}
\end{table*}

\begin{figure*}
    \centering
    \subfigure[left eyebrow]{\includegraphics[width=0.16\textwidth]{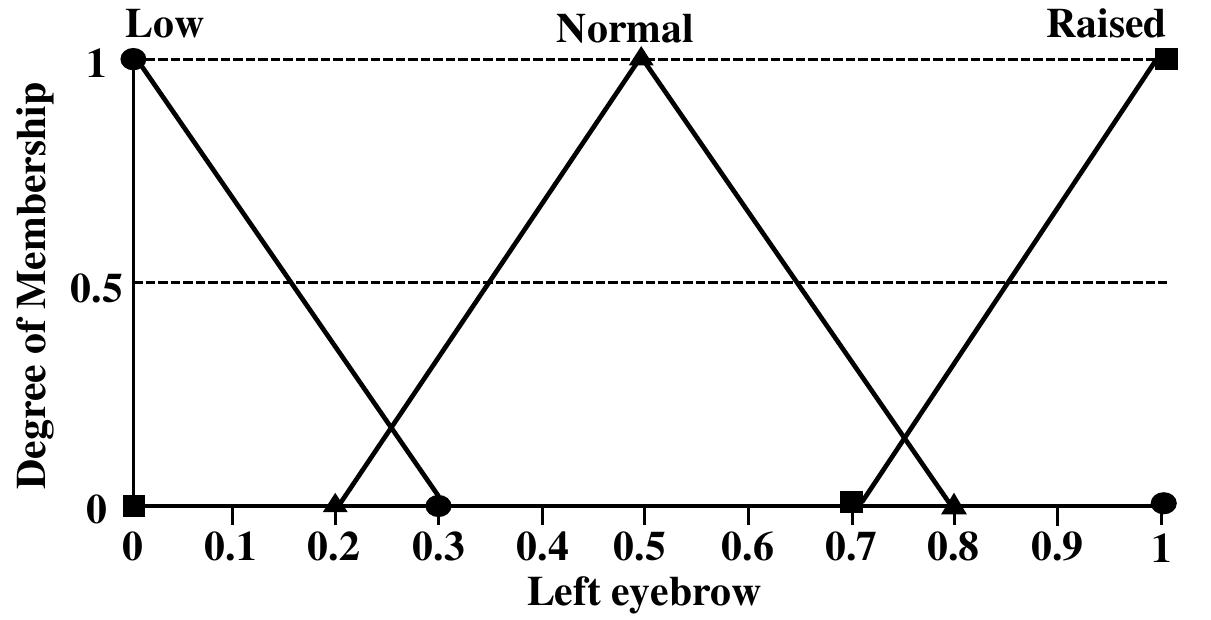}\label{fig:fcis_a}}
    \subfigure[right eyebrow]{\includegraphics[width=0.16\textwidth]{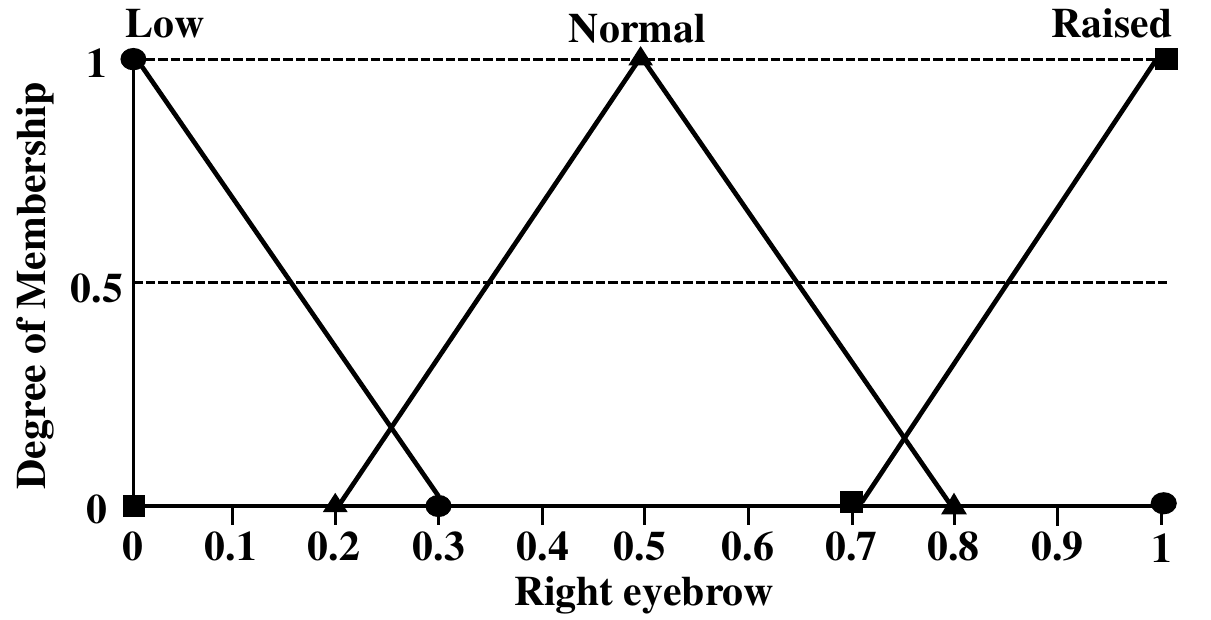}\label{fig:fcis_b}}
    \subfigure[brow crest]{\includegraphics[width=0.16\textwidth]{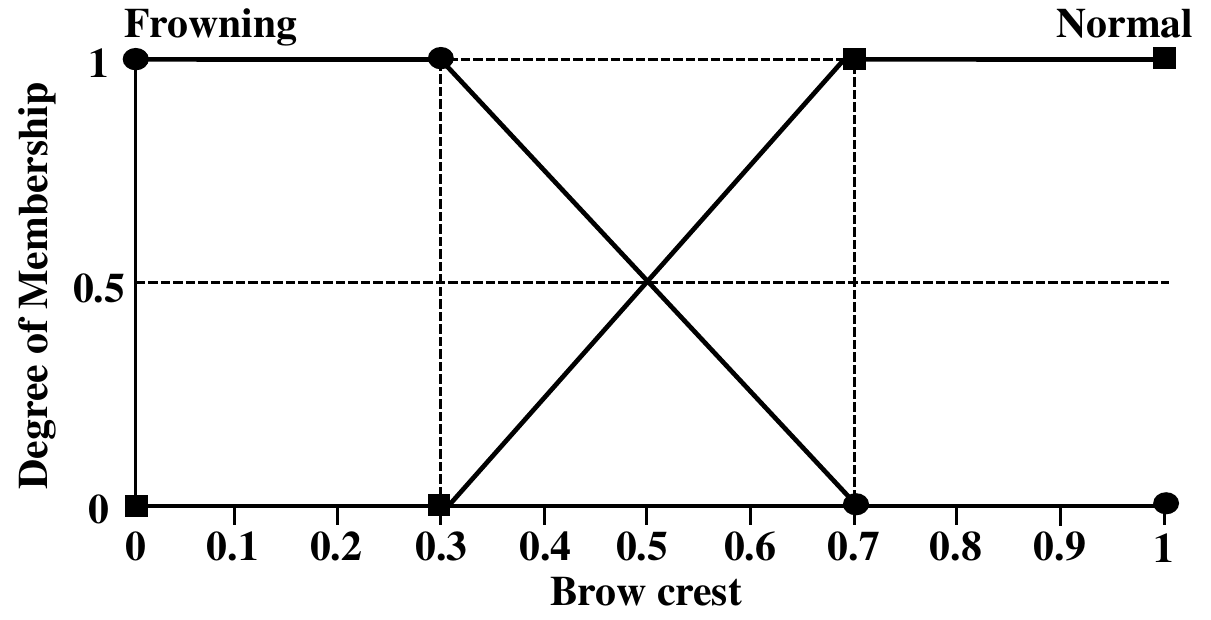}\label{fig:fcis_c}}
    \subfigure[left eye]{\includegraphics[width=0.16\textwidth]{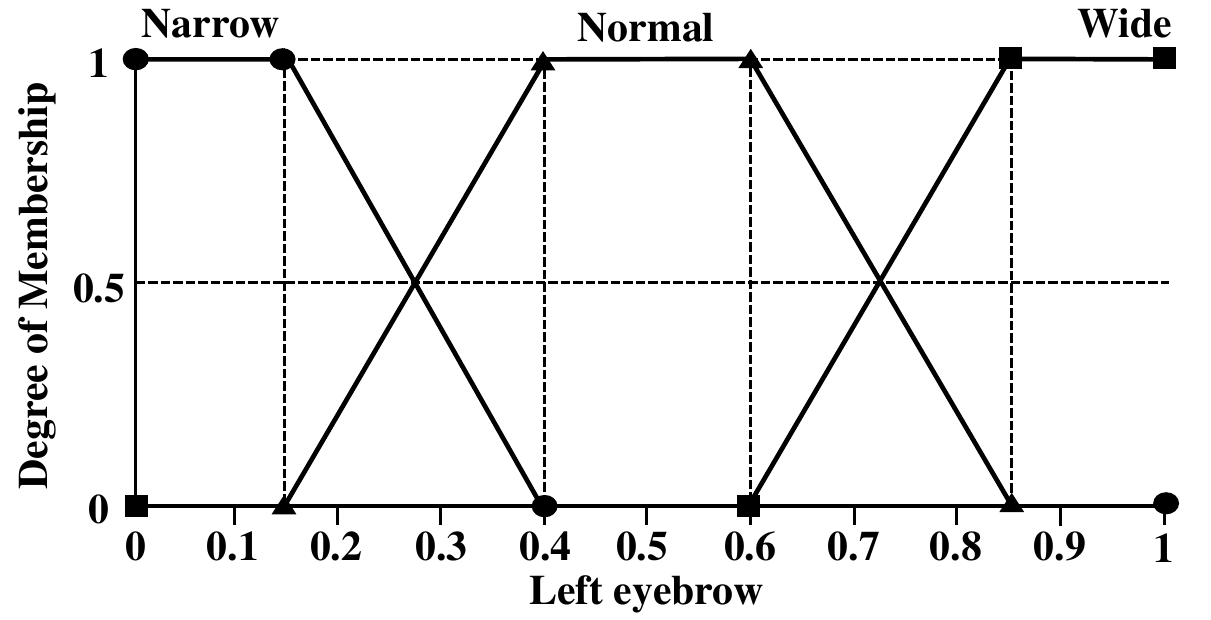}\label{fig:fcis_d}}
    \subfigure[right eye]{\includegraphics[width=0.16\textwidth]{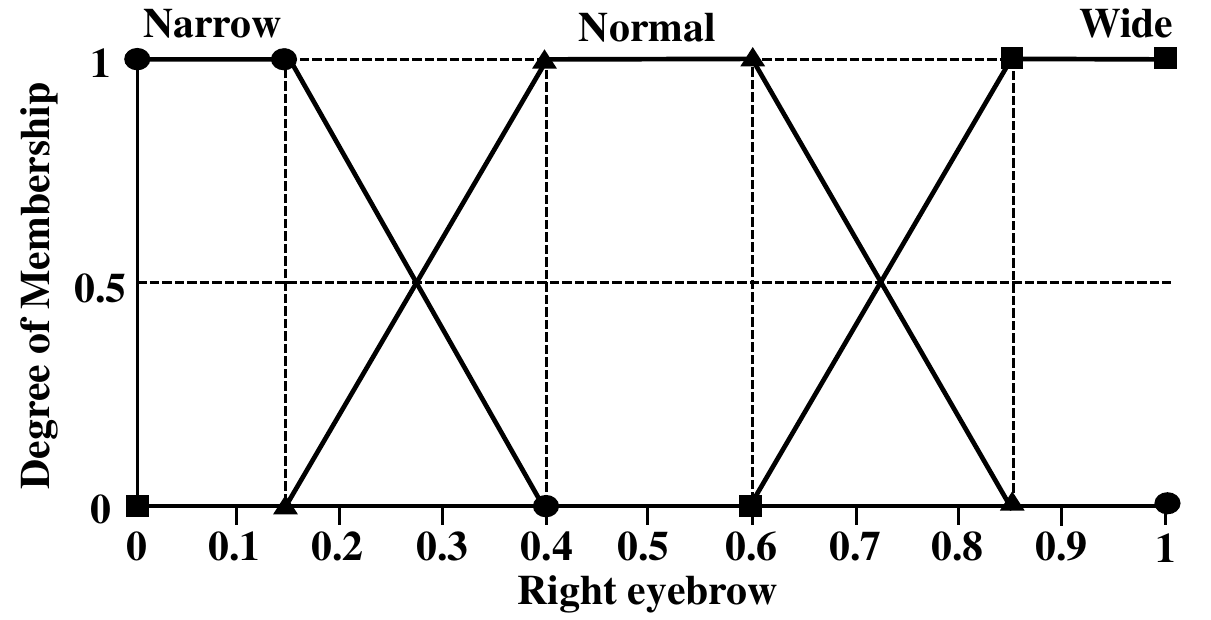}\label{fig:fcis_e}}
    \subfigure[nose]{\includegraphics[width=0.16\textwidth]{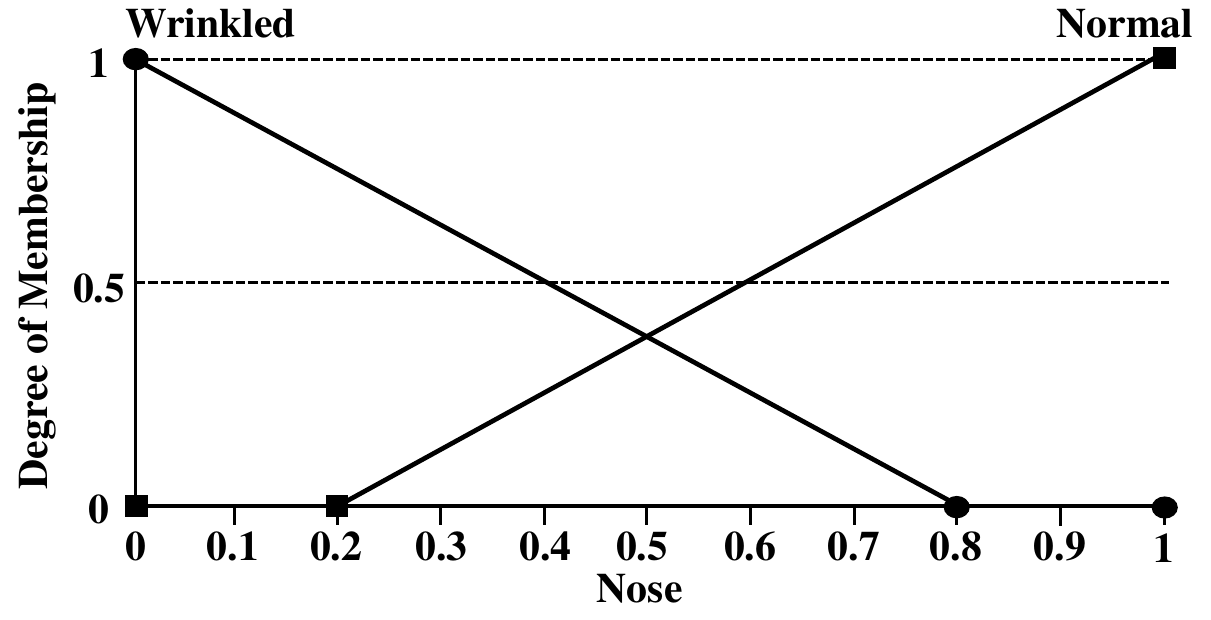}\label{fig:fcis_f}}
    \subfigure[nostril]{\includegraphics[width=0.16\textwidth]{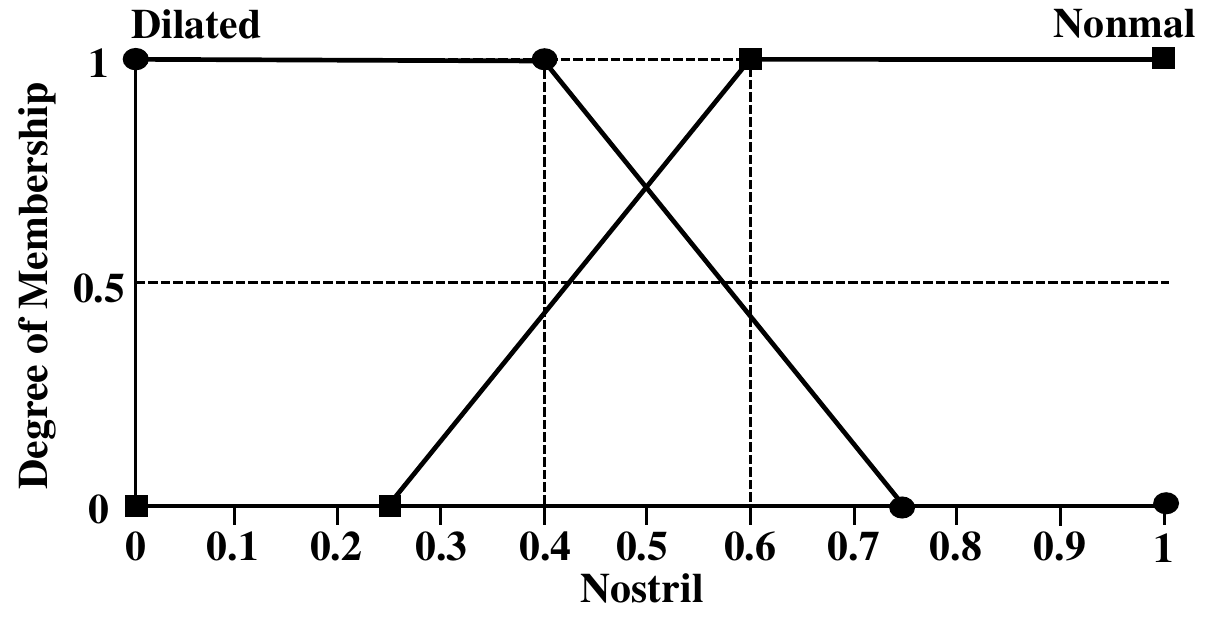}\label{fig:fcis_g}}
    \subfigure[mouth]{\includegraphics[width=0.16\textwidth]{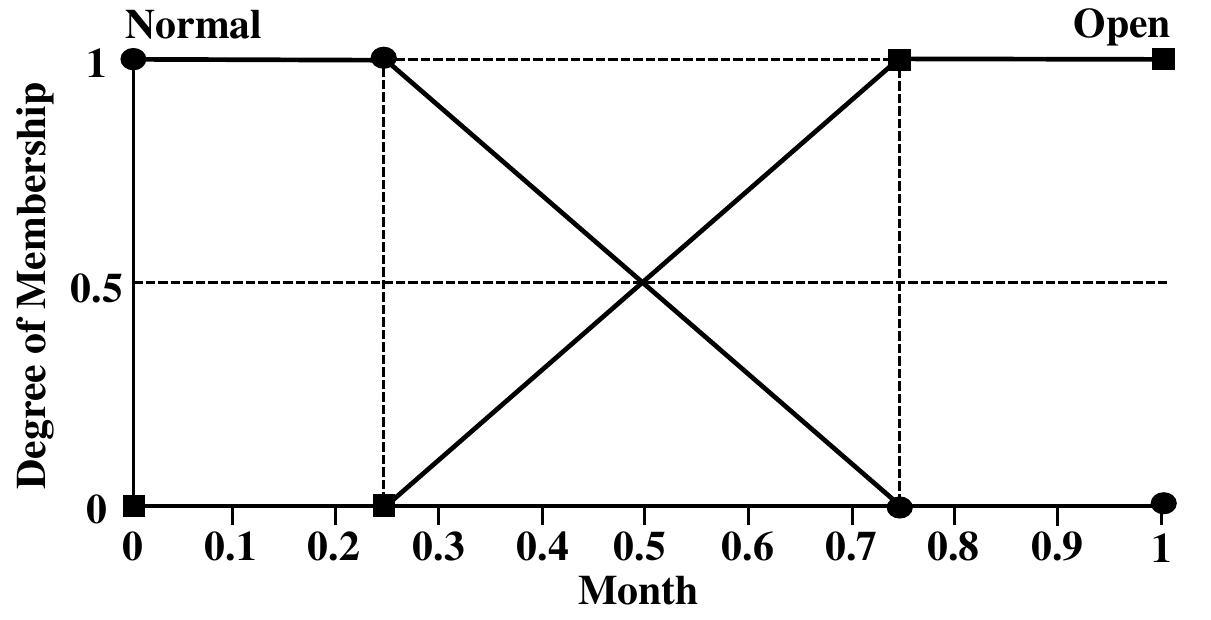}\label{fig:fcis_h}}
    \subfigure[upper lip]{\includegraphics[width=0.16\textwidth]{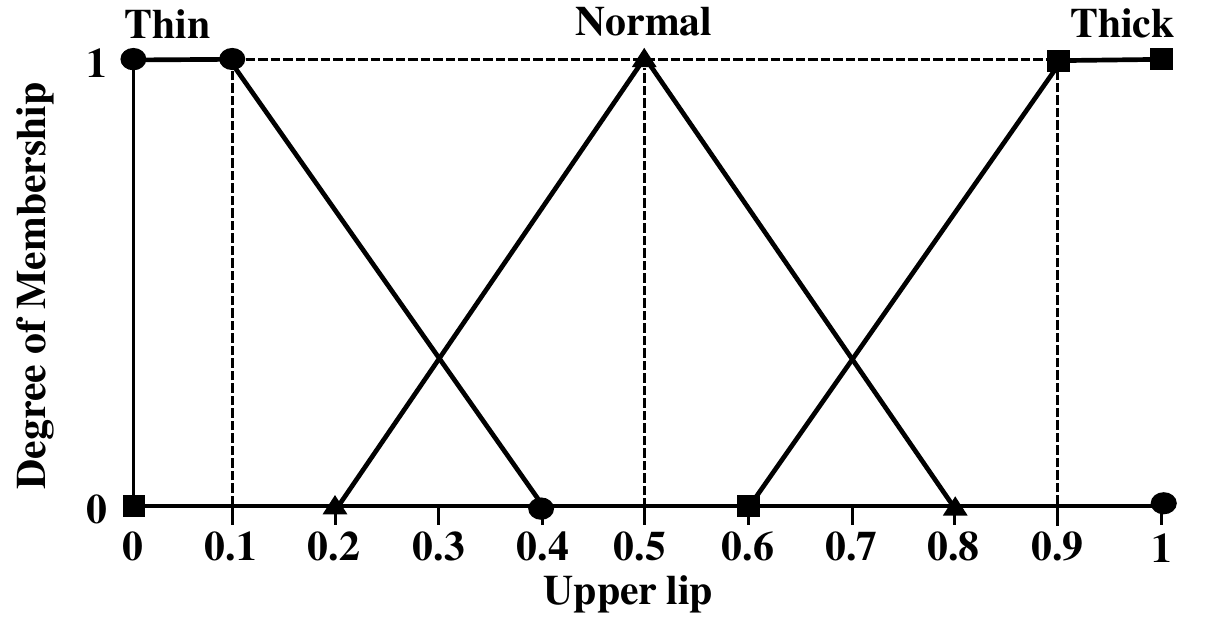}\label{fig:fcis_i}}
    \subfigure[lower lip]{\includegraphics[width=0.16\textwidth]{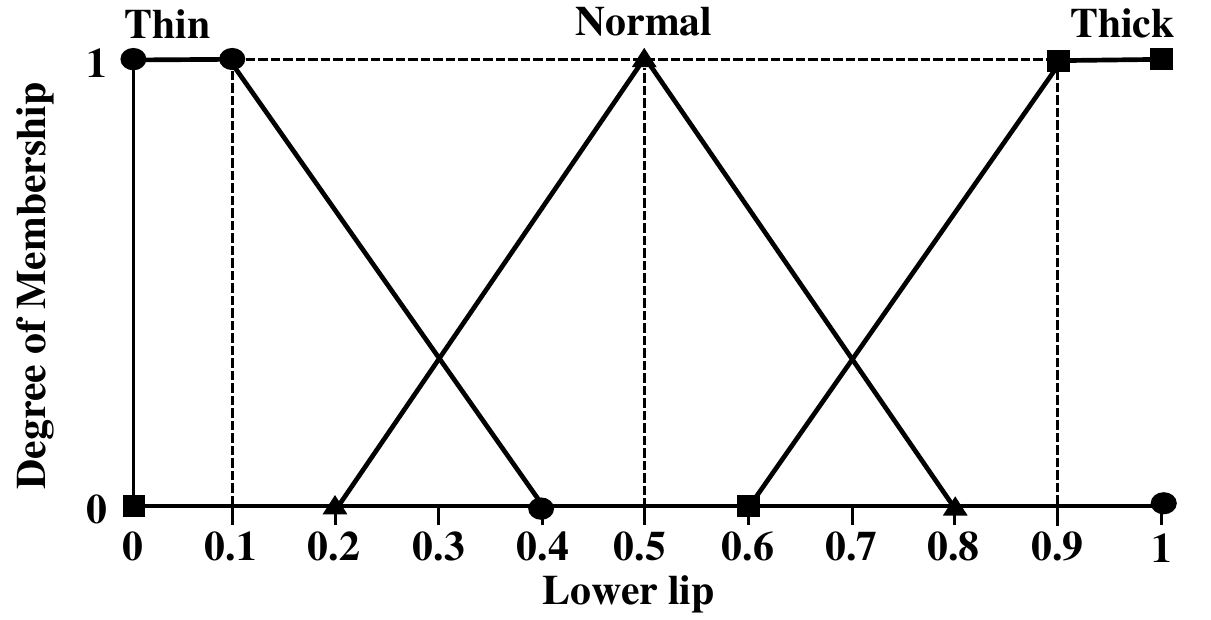}\label{fig:fcis_j}}
    \subfigure[left mouth corner]{\includegraphics[width=0.16\textwidth]{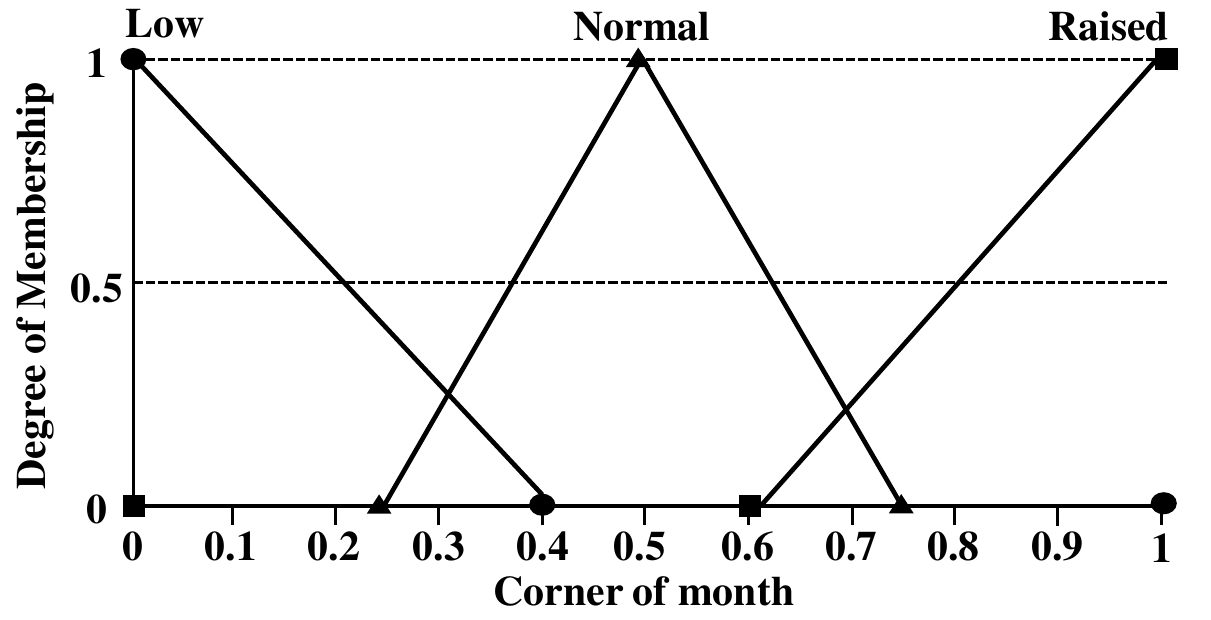}\label{fig:fcis_k}}
    \subfigure[right mouth corner]{\includegraphics[width=0.16\textwidth]{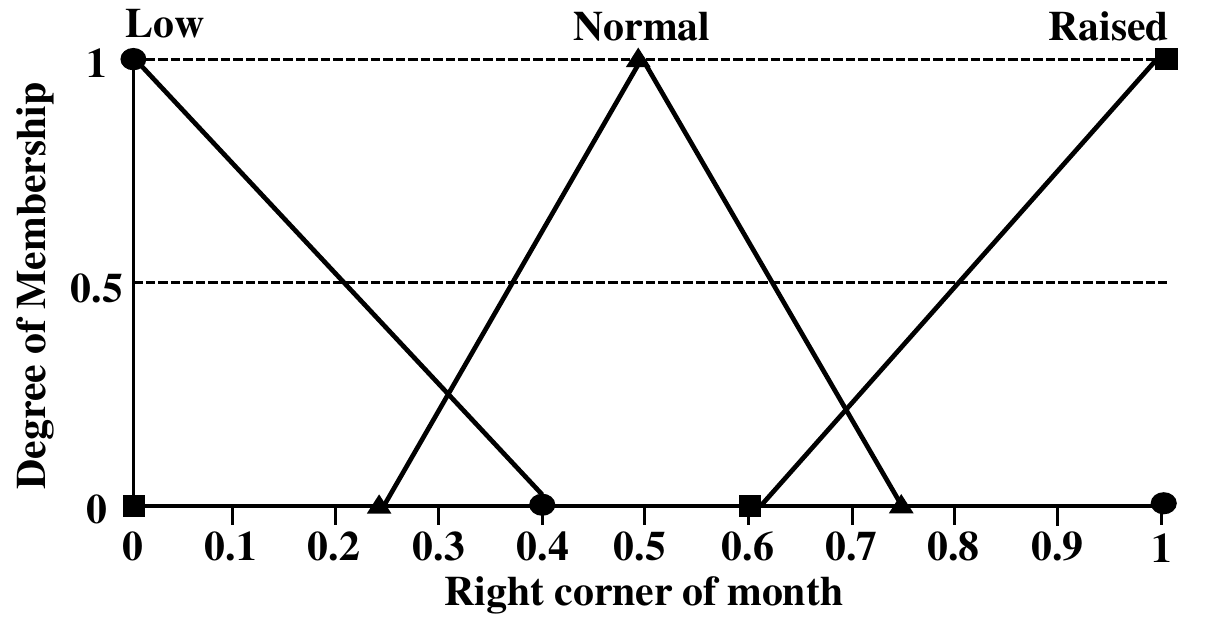}\label{fig:fcis_l}}
    \caption{The membership functions for all the physiological attributes. The horizontal axis represents the input, such as the feature encoding result of the neural network, and the vertical axis represents the membership degree of belonging to a certain attribute value. Taking (a) as an example, the three functions in the subfigure respectively represent the probability that the facial component left eyebrow belongs to low, normal, and raised.}
    \label{fig:2.1}
\end{figure*}

\begin{figure*}
    \centering
    \subfigure[Angry]{\includegraphics[width=0.16\textwidth]{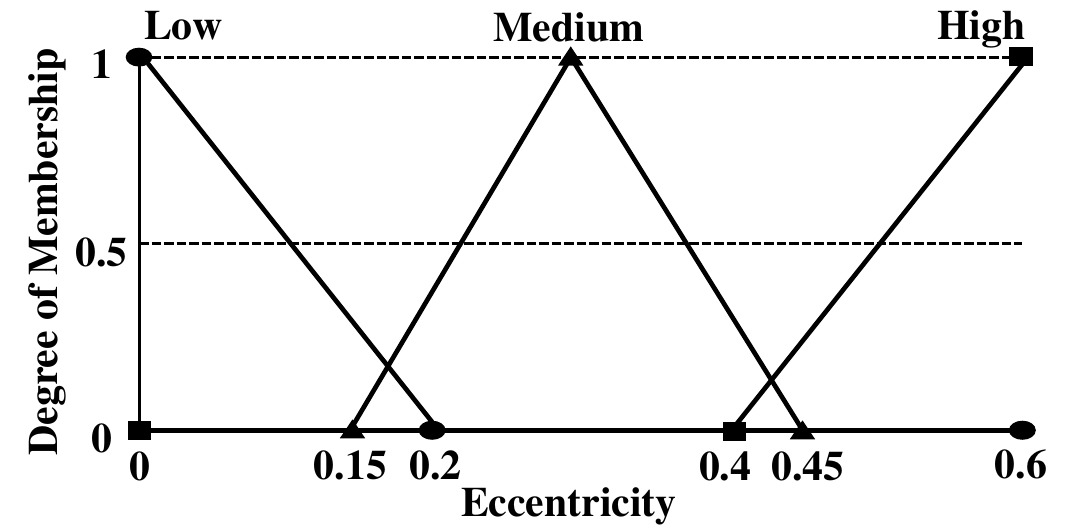}\label{fig:fkis_a}}
    \subfigure[Happy]{\includegraphics[width=0.16\textwidth]{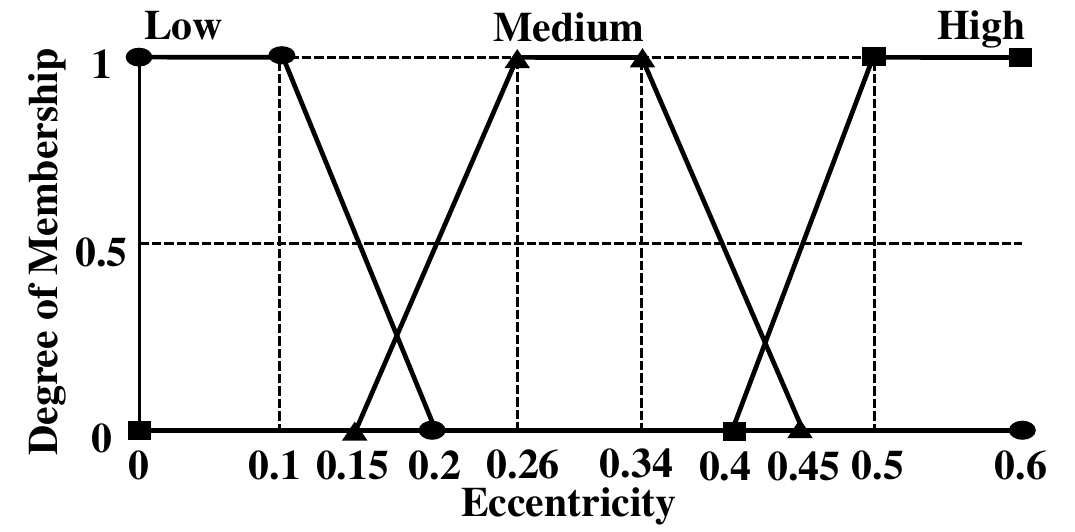}\label{fig:fkis_b}}
    \subfigure[Disgust]{\includegraphics[width=0.16\textwidth]{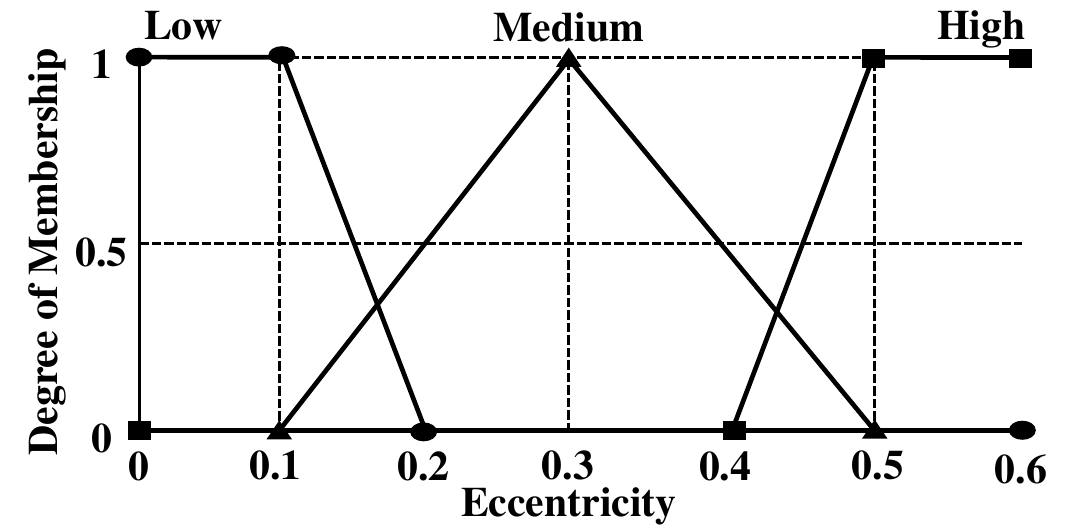}\label{fig:fkis_c}}
    \subfigure[Fear]{\includegraphics[width=0.16\textwidth]{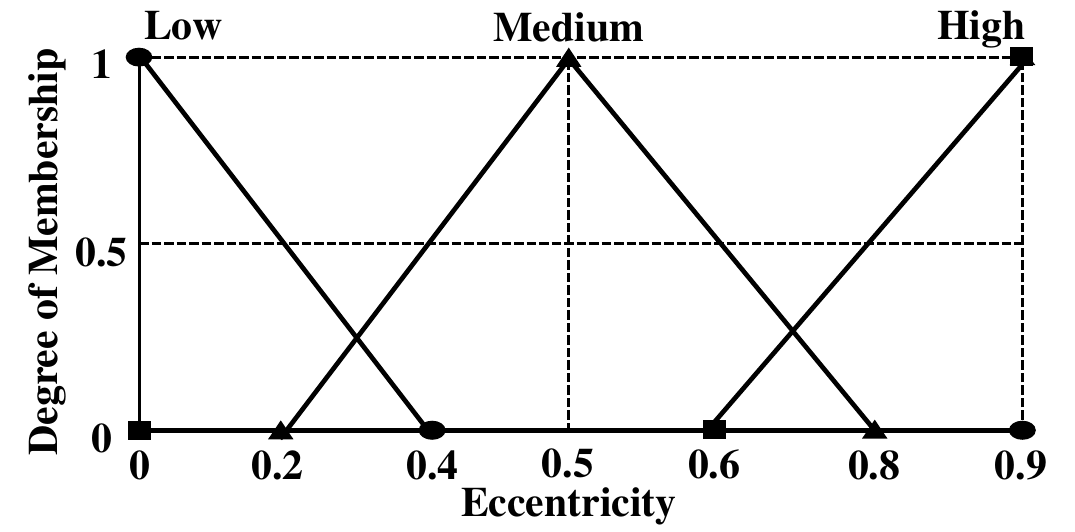}\label{fig:fkis_d}}
    \subfigure[Sad]{\includegraphics[width=0.16\textwidth]{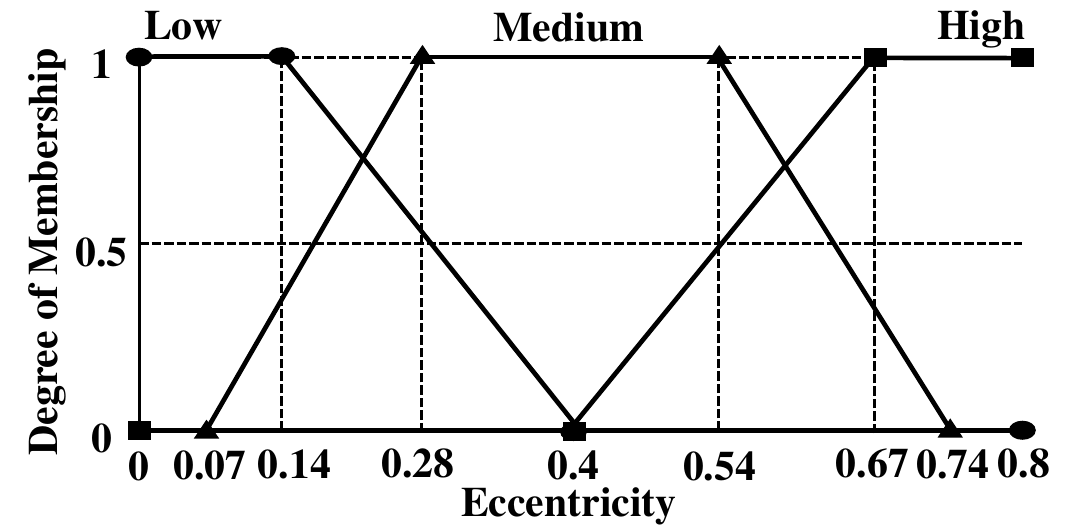}\label{fig:fkis_e}}
    \subfigure[Surprise]{\includegraphics[width=0.16\textwidth]{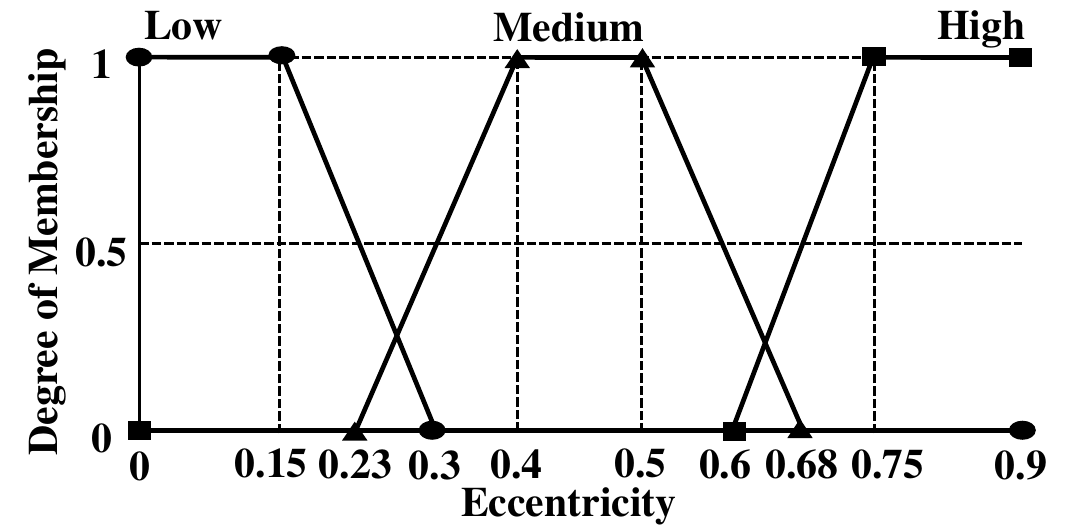}\label{fig:fkis_f}}
    \caption{The membership functions example for the intensity value of six emotions. The horizontal axis represents the converted eccentricity value of the coding of $fc_i$, while the vertical axis represents the membership degree that the coding result belongs to the corresponding emotional intensity. Each emotion contains three intensity values, i.e., low, medium, and high, which are calculated by three functions corresponding to the subfigure.}
    \label{fig:2.2}
\end{figure*}

\section{Generalized Fuzzy-based Rules}
\label{sec:3.2}
To quantify emotional intensity, we propose a set of generalized fuzzy-based rules to describe emotional states. By incorporating fuzzy inference into machine learning, we are better equipped to handle the inherent uncertainty in emotions, while also achieving higher computational efficiency in few-shot tasks \cite{yang2023cluster}. We develop two systems based on these proposed rules: fuzzy component inference system (FCIS), and fuzzy knowledge inference system (FKIS). 

\textbf{FCIS} aims to convert representations into a set of fuzzy linguistic conditions. Each condition describes a sequence of changes in a physiological attribute, where the elements of the sequence represent the state of that attribute. For instance, facial components have multiple attributes like eyebrows and mouth corners. These components change differently when humans express different emotions. Taking the eyebrow as an example, the values corresponding to low, normal, and raised are respectively $-1$, $0$, and $1$. When describing sad, the value of the eyebrows is $-1$, i.e., low, and when happy, the value of the eyebrows is $1$, i.e., raised. A linguistic condition describing a sequence of changes in the mouth could be represented as $(-1)(-1)(-1)011000$. Table \ref{tab:1} presents various attributes of facial components, while Figure \ref{fig:2.1} provides the membership functions for all the physiological attributes. The conditions generated by FCIS are then passed to FKIS.

\textbf{FKIS} aims to obtain sequences describing emotional intensity. The fuzzy inference rules within FKIS are constructed based on psychological knowledge \cite{koole2009psychology, kaur2012comparison}, and stored in a knowledge-based engine. The engine takes the changes in a set of physiological attributes as input and computes fuzzy sets based on the membership functions for each attribute. Subsequently, the engine generates 18 classification values for (composite) emotional mappings by combining trigger rules and performing de-fuzzification based on the fuzzy sets. Each basic emotion has three intensity levels, i.e., low, medium, and high. Table \ref{tab:2} provides a set of FKIS rules for determining emotional intensity, with each element representing the current state of a physiological attribute. Figure \ref{fig:2.2} illustrates the membership functions for the intensity value of six base emotions. Take Figure \ref{fig:2.2} (a) as an example, when ``Eccentricity=0.2'', the membership degree of ``Angry-Low'' is 0, and the membership degree of ``Angry-Medium'' is 0.34. The coding of the emotions refers to representing emotions as numerical or symbolic values, while membership functions are used to describe the degree of belonging of each data point to each fuzzy set, i.e., the extent of each data point in each emotion category, or its degree in each facial component. We weight the emotional intensity encoding (shown in Table \ref{tab:2}) based on the membership value and introduce it into the model calculation as fuzzy information. The fuzzy membership functions calculate the coding of the emotions for each view and introduce one-dimensional information with fuzzy semantics, which better captures the complexity of emotions. Note that different emotions have varying numbers of rules; for instance, anger has 42 rules, while disgust only has 34 rules.

\begin{figure*}[t]
    \centering
    \includegraphics[width=\textwidth]{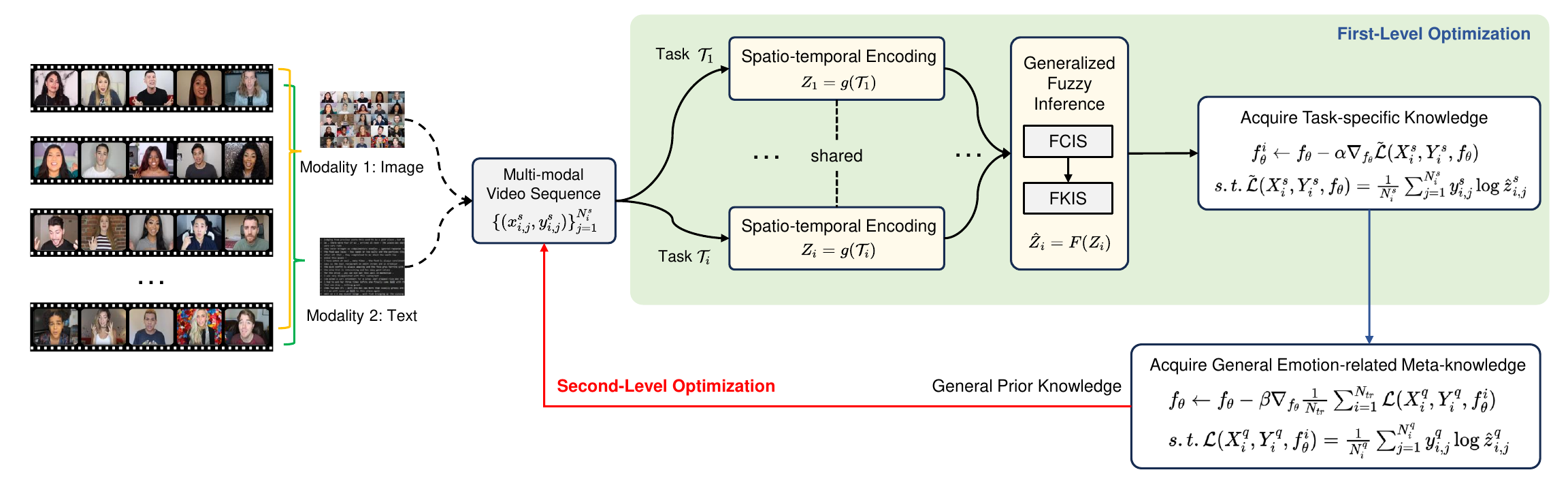}
    \caption{The framework of ST-F2M. ST-F2M first constructs multi-modal meta-learning tasks based on the input videos. Next, it performs encoding through a spatio-temporal convolutional integrated module, and then uses generalized fuzzy inference to add fuzzy emotional semantic information. Finally, ST-F2M acquires task-specific and general FER knowledge through bi-level optimization to achieve fast and robust FER. The pseudo-code is shown in Algorithm \ref{alg:algorithm}.}
    \label{fig:st-f2m}
\end{figure*}

In this study, the above systems, i.e., FCIS and FKIS, are used in two goals: (i) for the decision-making process of ST-F2M: followed by the spatio-temporal integrated module to assign fuzzy semantic information containing emotional intensity to the representation; (ii) for dataset construction: assign emotion and intensity information to benchmark datasets in areas such as facial expression recognition to build benchmark datasets for fine-grained emotion recognition. 

For the first objective, the introduction of fuzzy rules helps deal with the complexity and ambiguity of emotions in reality and improve model generalization. Specifically, the proposed emotional reasoning system is grounded in psychological theories of human affect, wherein complex emotional states are decomposed into a set of elementary characteristic components. These components serve as the minimal semantic units of emotion, analogous to ``building blocks'', and enable a compositional representation of emotional expressions. This formulation allows emotional data, regardless of context or modality, to be interpreted as permutations of these basic units, thus supporting generalizability and reusability across different scenarios. Further, the fuzzy membership mechanism allows each component to be activated with varying intensities depending on context and subject, addressing the intrinsic individual variability and ambiguity in emotional expression. This enables the system to model the uncertainty and continuity inherent in human emotions more robustly than discrete categorical methods. The specific implementation is illustrated in Section \ref{sec:4}.

For the second objective,  our system significantly reduces data maintenance overhead compared with traditional annotation-based approaches, which demand costly manual labeling for every new video or sample. First, both building a fuzzy inference system and data annotation for the first time require human knowledge as a prior. However, the systems we built are based on fuzzy emotional reasoning rules constructed from human characteristic components, which are universal and applicable to all emotions once and for all. For new data, the features can still be extracted by the feature extractor and then input into the inference system for fuzzy inference without the need to reconstruct the rules. In contrast, data annotation is to identify and annotate videos from the pixel level. New video data needs to be re-annotated, which will incur new costs. In addition, our fuzzy system contains a built-in inference function. We only need to write down the encoding ${-1,0,1}$ of the characteristic components required for the emotion, and then we can perform inference based on the built-in functions in FCIS and FKIS without requiring a separate model training process. Thus, we can generate new fuzzy rules based on the inference function at any time without additional cost. Therefore, the introduction of fuzzy rules greatly improves the model effect and speeds up reasoning, while the cost of first construction is negligible for the model. We have constructed two FER datasets, called DISFA-FER and WFLW-FER, based on DISFA \cite{mavadati2013disfa} and WFLW \cite{wu2018look}.

\section{Methodology}
\label{sec:4}
To address the three key challenges in FER, we propose a spatio-temporal fuzzy-oriented multi-modal meta-learning framework (ST-F2M). During the design process of the framework, we aim to solve four key issues: (i) how to construct multi-modal meta-learning tasks, (ii) how to consider spatio-temporal heterogeneity, (iii) how to deal with the complexity and ambiguity of emotions, and (iv) how to learn general prior knowledge from limited data and achieve rapid convergence. In this section, we first describe how to partition the limited training data into several meta-learning tasks and then provide details about the model training process. The overall framework of ST-F2M is illustrated in Figure \ref{fig:st-f2m}, and the pseudo-code of ST-F2M is shown in Algorithm \ref{alg:algorithm}.

\subsection{Meta-learning Task Construction}
\label{sec:4.1}

Before implementing meta-learning, we first build meta-learning tasks for fine-grained emotion recognition. Our goal in this subsection is to solve the first issue mentioned above, i.e., how to construct multi-modal meta-learning tasks.

Specifically, we first separate different modalities from the input long video data and add location labels to each modal data. The location label refers to the relative position (appearance order) between different elements of each modality. It is obtained by the positional encoding of different modal data in the long video through an MLP, without requiring an additional manual annotation process. Next, we segment the data of different modalities according to the emotional label sequences, and each segment corresponds to an emotional view. Each view reflects a modal data of an emotion. Then, views from the same location are combined into a group based on the location labels. Finally, we randomly select multiple groups of views with the same emotion labels to form a meta-learning task. Each modal data for each emotion corresponds to a perspective.

Formally, given the FER videos $(\mathcal{X},\mathcal{Y})$ where $\mathcal{X}_a$ and $\mathcal{Y}_a$ are the $a$-th long video and the corresponding label sequence respectively, we first segment the visual modality and text modality data represented as $(\mathcal{X}^v_a,\mathcal{Y}_a,pos_a)$ and $(\mathcal{X}^t_a, \mathcal{ Y}_a,pos_a)$, where $pos_i$ is the positional encoding of different modal data in the long video. Next, we segment the data based on the emotional label sequence $\mathcal{Y}_a$ and obtain $\mathcal{X}^v_a=\left \{ x_{a,b}^v \right \}_{b=1 }^{N_{\mathcal{Y}_a}}$ and $\mathcal{X}^t_a=\left \{ x_{a,b}^t \right \}_{b=1}^{N_{ \mathcal{Y}_a}}$, where $N_{\mathcal{Y}_a}$ is the number of emotions contained in the long video, that is, the number of elements in $\mathcal { Y}_a$.  Next, we aggregate views with the same position to obtain $x_{a,b}=\left \{ x_{a,b}^v, x_{a,b}^t\right \}$. Finally, a meta-learning task $\mathcal{T}_i$ is formed by randomly selecting $x_{a,b}$ with the same emotional label.

\subsection{Meta-learning Training}
\label{sec:4.2}

Next, we will perform meta-learning based on the constructed tasks mentioned above. Our goal in this section is to address the last three issues mentioned above, and propose a spatio-temporal convolution integrated module, a generalized fuzzy inference module, and a meta-optimation module. Next, we will introduce the training process of ST-F2M, as well as the details of these three modules.

\subsubsection{\textbf{Overview of meta-training}}
The training process of ST-F2M is mainly divided into four steps: (i) encoding through the spatio-temporal convolution integrated module, which helps consider spatio-temporal heterogeneity; (ii) adding fuzzy emotional semantic information using generalized fuzzy inference, which helps deal with the complexity
and ambiguity of emotions; (iii) emotion category determination based on classification head, which obtains FER results; and (iv) bi-level optimization through the meta-optimation module, which improves model efficiency and robustness. Through this process, ST-F2M can effectively cope with the three challenges of real-life FER and achieve rapid convergence from small and complex multi-modal emotion data.

Formally, ST-F2M can be defined as $f_{\theta}=h\circ F\circ g$, where $g$ is the shared encoder, $F$ denotes the fuzzy inference, $h$ is the classifier. We first construct the training set $\mathcal{D}_{tr}$ and the test set $\mathcal{D}_{te}$ based on the method in Subsection \ref{sec:4.1}, which are obtained from the distribution $p(\mathcal{T})$. For each task $\mathcal{T}_i$, we first input it into the encoder $g$ composed of a spatio-temporal convolution integrated module, thereby obtaining the representation $Z_i=g(\mathcal{T}_i )$. Next, $Z_i$ is passed into the fuzzy inference system $F$ to obtain the representation $\hat{Z}_i=F(Z_i)$ that introduces fuzzy semantic information. Then, $\hat{Z}_i$ will be input into the classification header to generate specific classification results. Ultimately, this process will undergo a bi-level optimization as mentioned in Subsection \ref{sec:3} to learn common meta-knowledge for rapid adaptation to new tasks.

\begin{algorithm}[t]
\caption{Pseudo-Code of ST-F2M}
\label{alg:algorithm}
    \textbf{Input}: Task distribution $p(\mathcal{T})$; Randomly initialize model $f_{\theta}$ with an encoder $g(\cdot)$, a fuzzy inference module $F$, and a classifier $h(\cdot)$; FCIS and FKIS\;
    \textbf{Parameter}: Mini-batch $B$; Learning rates $\alpha$ and $\beta$\;
    \textbf{Output}: The trained model $f_\theta$\;
  \tcc{TASK CONSTRUCTION}
  \For{each task}{
  Sample a mini-batch $\mathcal{B}_{\boldsymbol{x}}=(\mathcal{X},\mathcal{Y})$ from $p(\mathcal{T})$\;
  Calculate the positional encoding $pos_i$ for $(\mathcal{X},\mathcal{Y})$\;
  Segment $(\mathcal{X},\mathcal{Y})$ into the visual modality $(\mathcal{X}^v_a,\mathcal{Y}_a,pos_a)$ and text modality $(\mathcal{X}^t_a, \mathcal{ Y}_a,pos_a)$\;
  Segment $(\mathcal{X}^v_a,\mathcal{Y}_a,pos_a)$ based on the emotional label sequence $\mathcal{Y}_a$, obtaining $\mathcal{X}^v_a=\left \{ x_{a,b}^v \right \}_{b=1 }^{N_{\mathcal{Y}_a}}$\; 
  Segment $(\mathcal{X}^t_a, \mathcal{ Y}_a,pos_a)$ based on the emotional label sequence $\mathcal{Y}_a$, obtaining $\mathcal{X}^t_a=\left \{ x_{a,b}^t \right \}_{b=1}^{N_{ \mathcal{Y}_a}}$\;
  Randomly select $x_{a,b}$ with the same emotional label, obtaining task $\mathcal{T}_i$\; 
  }
  \tcc{META-LEARNING}
  \For{$n$ tasks}{
  \tcc{BI-LEVEL OPTIMIZATION}
    \For{each $\mathcal{T}_i$ with the dataset $\mathcal{D}_{i}$}{ 
    \tcc{FIRST-LEVEL}
    Divide the data $\mathcal{D}_{i}$ into the support set $\mathcal{D}_{i}^s$ and the query set $\mathcal{D}_{i}^q$\;
    Encode $\mathcal{T}_i$ using spatio-temporal convolution integrated module via Eq.\ref{eq:TC} and Eq.\ref{eq:C_sp}, obtaining $Z_i=g(\mathcal{T}_i)$\;
    Perform generalized fuzzy inference with FCIS and FKIS mentioned in Subsection \ref{sec:3.2}, obtaining $\hat{Z}_i=F(Z_i)$\;
    Update the task-specific model $f^i_{\theta}$ using Eq.\ref{eq:inner_st-f2m}, i.e., $f^i_{\theta} \gets f_{\theta} -\alpha \nabla_{f_{\theta}}\tilde{\mathcal{L}} (X_i^s,Y_i^s,f_{\theta})$\;
    }
    \tcc{SECOND-LEVEL}
    Update the ST-F2M model $f_\theta$ using Eq.\ref{eq:outer_st-f2m}, i.e., $f_{\theta} \gets f_{\theta}-\beta \nabla_{f_{\theta}}\frac{1}{N_{tr}}\sum_{i=1}^{N_{tr}}\mathcal{L}(X_i^q,Y_i^q,f_\theta ^i)$ \
   }
\end{algorithm}

\subsubsection{\textbf{Spatio-temporal encoding}}

We first introduce the process of spatio-temporal encoding, i.e., $Z_i=g(\mathcal{T}_i )$. 

To capture the temporal correlation, we adopt a one-dimensional convolution following \cite{yu2017spatio} along the temporal dimension with a gating mechanism. Specifically, our temporal convolution ($C_{te}$) takes the task $\mathcal{T}_i$ composed of FER video tensors as input, while outputting a time-aware embedding of each emotional state:
\begin{equation}
\label{eq:TC}
    (\mathcal{Z} ^{t-t_{out}}_i,...,\mathcal{Z} ^{t}_i)=C_{te}(\mathcal{T}_i )
\end{equation}
where $t_{out}$ denotes the length of the output embedding following the convolutional operations in the encoder, $\mathcal{Z} ^{t}_i\in \mathbb{R}^{N_s\times d} $ denotes the embedding matrix at the time step $t$, $N_s$ and $d$ represent the number of samples in the support set and embedding dimensions. The $n$-th row $z_i^{t,n}\in \mathbb{R}^d $ corresponds to the embedding of view $x_{i,j}^s$. Further, in order to capture the spatial correlation, we construct a spatial convolution module ($C_{sp}$) based on the message-passing mechanism \cite{cai2021rethinking}, and integrate the temporal information:
\begin{equation}
\label{eq:C_sp}
    Z_i^t=C_{sp}(\mathcal{Z}_i^t,A_i )
\end{equation}
where $A_i$ is the region adjacency matrix of task $\mathcal{T}_i$, $Z_i$ denotes the refined embedding of task $\mathcal{T}_i$ which takes into account spatio-temporal heterogeneity. This spatio-temporal encoder effectively captures emotional changes in long videos, allowing us to model the connection between emotional states across time and various spatial regions, which solves the second issue.

\subsubsection{\textbf{Generalized fuzzy inference}}

For generalized fuzzy inference, we perform secondary encoding based on the fuzzy systems described in Subsection \ref{sec:3.2}, that is, $\hat{Z}_i=F(Z_i)$. 

In brief, after spatio-temporal encoding, the feature embedding $Z_i$ is input into FCIS, which directly calculates the values of $fc_i$ based on the designed membership function (as shown in Table \ref{tab:1} and Figure \ref{fig:2.1}). Next, $Z_i$ is input into FKIS, which assigns a one-dimensional fuzzy semantics to $\hat{Z}_i$ based on the designed membership function of FKIS (as shown in Table \ref{tab:2} and Figure \ref{fig:2.2}) and the result of FCIS. The optimization of the generalized fuzzy inference module is performed when building the inference system before meta-training, and its fitness function \cite{nelson2009fitness, baresel2002fitness} is expressed as:
\begin{equation}
\label{eq:loss_fuzzy}
    \mathcal{F}_ F=\lambda_1 \mathcal{F}_ {FCIS} +\lambda_2 \mathcal{F}_ {FKIS}
\end{equation}
where $\lambda_1$ and $\lambda_2$ represent the range of membership functions in FCIS and FKIS, which affects the final meta-learning model performance. The calculation of this loss is obtained by evaluating the final effect of changing the fuzzy rules through a fixed meta-learning model.

Compared with $Z_i$, one-dimensional information containing fuzzy semantics is introduced into $\hat{Z} _i$, and its element value is the encoding information of $fc_i$. Next, the embedding $\hat{Z} _i$ containing fuzzy information of each view will be input to the classifier $h$ for fine-grained emotion recognition to obtain the emotion change sequence of task $\mathcal{T}_i$. The generalized fuzzy inference module we built can give the data fuzzy semantic information to cope with the differences in emotional expressions between different individuals and the similarities between different emotions. It addresses the third issue regarding emotional complexity and ambiguity.

\subsubsection{\textbf{Overall optimization}}

Finally, we utilize meta-recurrent neural networks to capture general emotion-related meta-knowledge and achieve fast convergence for FER, i.e., updating the model $f_{\theta}=h\circ F\circ g$.

Specifically, we perform bi-level optimization as described in Subsection \ref{sec:3}. The objective of the first level optimization (inner loop) is updated to:
\begin{equation}
\begin{array}{l}
\label{eq:inner_st-f2m}
    \qquad f^i_{\theta} \gets f_{\theta} -\alpha \nabla_{f_{\theta}}\tilde{\mathcal{L}} (X_i^s,Y_i^s,f_{\theta}) \\[8pt]
    s.t. \quad \tilde{\mathcal{L}}(X_i^s,Y_i^s,f_{\theta} )=\frac{1}{N_i^s} \sum_{j=1}^{N_i^s}y_{i,j}^s\log \hat{z} _{i,j}^{s}
\end{array}
\end{equation}
where $\alpha$ is the learning rate, $N_i^s$ is the number of samples in the support set $\mathcal{D}_i^s $, $\hat{z} _{i,j}^{s}\in \hat{Z} _i^{s}$ is the embedding of the view $x_{i,j}^s$ calculated through $g$, $F$, and $h$.

Then, the objective of the second level (outer loop) becomes:
\begin{equation}
\begin{array}{l}
\label{eq:outer_st-f2m}
    f_{\theta} \gets f_{\theta}-\beta \nabla_{f_{\theta}}\frac{1}{N_{tr}}\sum_{i=1}^{N_{tr}}\mathcal{L}(X_i^q,Y_i^q,f_\theta ^i)\\[8pt]
    s.t. \quad \mathcal{L}(X_i^q,Y_i^q,f_\theta ^i)= \frac{1}{N_i^q} \sum_{j=1}^{N_i^q}y_{i,j}^q\log \hat{z} _{i,j}^q
\end{array}
\end{equation}
where $\beta$ is the learning rate, $N_i^q$ represents the number of samples in the query set $\mathcal{D}_i^q$, $N_{tr}$ is the number of training tasks, $\hat{z} _{i,j}^{q}\in \hat{Z} _i^{q}$ denotes the embedding of the view $x_{i,j}^q$ in the query set $\mathcal{D}_i^q$, which is calculated through $g$, $F$, and $h$.

Through this bi-level optimization mechanism, we can learn emotion-related general meta-knowledge from training tasks and quickly adapt to new tasks based on this knowledge. This meta-optimization module solves the fourth issue in the real-life FER problem mentioned above, which is to learn key knowledge from limited data and achieve rapid convergence.

\begin{table*}
  \centering
  \caption{Comparison with Baseline Models on Five Benchmark Datasets. The values in the table represent the Accuracy(\%) and Recall(\%) with a 95\% confidence interval, and the optimal values will be highlighted in bold.}
  \label{tab:ex1}
  \resizebox{\linewidth}{!}{
  \begin{tabular}{l|cc|cc|cc|cc|cc}
    \toprule
    \multirow{2}{*}{\textbf{Methods}} & \multicolumn{2}{c|}{\textbf{CK+}} & \multicolumn{2}{c|}{\textbf{DISFA-FER}} & \multicolumn{2}{c|}{\textbf{WFLW-FER}} & \multicolumn{2}{c|}{\textbf{CMU-MOSEI}} & \multicolumn{2}{c}{\textbf{CREMA-D}} \\
    & \textbf{ACC} & \textbf{RECALL} & \textbf{ACC} & \textbf{RECALL} & \textbf{ACC} & \textbf{RECALL} & \textbf{ACC} & \textbf{RECALL} & \textbf{ACC} & \textbf{RECALL} \\
    \midrule
    \textbf{FN2EN}        & 96.13 $\pm$ 0.61 & 94.56 $\pm$ 0.78 & 81.12 $\pm$ 0.75 & 84.35 $\pm$ 0.84 & 78.91 $\pm$ 0.24 & 80.85 $\pm$ 0.37 & 84.77 $\pm$ 0.20 & 85.09 $\pm$ 0.51 & 69.06 $\pm$ 0.26 & 71.87 $\pm$ 0.43  \\
    \textbf{VGG-F-CNN}    & 94.67 $\pm$ 0.23 & 95.01 $\pm$ 0.91 & 80.64 $\pm$ 0.58 & 86.19 $\pm$ 0.17 & 79.68 $\pm$ 0.28 & 80.39 $\pm$ 0.47 & 75.56 $\pm$ 0.73 & 75.94 $\pm$ 0.25 & 67.05 $\pm$ 0.12 & 69.08 $\pm$ 0.16  \\
    \textbf{PropNet-CNN}  & 95.07 $\pm$ 0.89 & 94.54 $\pm$ 0.07 & 77.51 $\pm$ 0.93 & 83.10 $\pm$ 0.42 & 81.80 $\pm$ 0.73 & 82.69 $\pm$ 0.57 & 72.21 $\pm$ 0.34 & 70.90 $\pm$ 0.46 & 66.71 $\pm$ 0.46 & 68.68 $\pm$ 0.54  \\
    \textbf{UniMSE}       & 89.28 $\pm$ 0.74 & 90.45 $\pm$ 0.13 & 79.57 $\pm$ 0.14 & 84.17 $\pm$ 0.55 & 80.20 $\pm$ 0.83 & 79.73 $\pm$ 0.12 & 86.97 $\pm$ 0.56 & 87.49 $\pm$ 0.36 & 64.43 $\pm$ 0.84 & 64.28 $\pm$ 0.06  \\
    \textbf{SepTr}        & 90.30 $\pm$ 0.47 & 88.31 $\pm$ 0.11 & 82.94 $\pm$ 0.27 & 82.97 $\pm$ 0.21 & 78.06 $\pm$ 0.77 & 77.54 $\pm$ 0.05 & 80.41 $\pm$ 0.50 & 81.67 $\pm$ 0.08 & 70.40 $\pm$ 0.64 & 71.81 $\pm$ 0.34  \\
    \midrule
    \textbf{MAML}         & 85.70 $\pm$ 0.64 & 87.51 $\pm$ 0.08 & 76.94 $\pm$ 0.30 & 77.07 $\pm$ 0.18 & 74.67 $\pm$ 0.29 & 77.83 $\pm$ 0.55 & 81.38 $\pm$ 0.72 & 81.64 $\pm$ 0.86 & 64.79 $\pm$ 0.68 & 65.50 $\pm$ 0.32  \\
    \textbf{Reptile}         & 85.15 $\pm$ 0.59 & 87.99 $\pm$ 0.30 & 77.56 $\pm$ 0.29 & 76.56 $\pm$ 0.35 & 75.51 $\pm$ 0.33 & 78.04 $\pm$ 0.40 & 83.51 $\pm$ 0.70 & 82.56 $\pm$ 0.56 & 63.41 $\pm$ 0.57 & 64.37 $\pm$ 0.25 \\  
    \textbf{ProtoNet}         & 86.40 $\pm$ 0.67 & 86.65 $\pm$ 0.43 & 78.43 $\pm$ 0.43 & 78.24 $\pm$ 0.63 & 73.76 $\pm$ 0.43 & 78.38 $\pm$ 0.53 & 83.43 $\pm$ 0.46 & 82.76 $\pm$ 0.45 & 65.35 $\pm$ 0.74 & 66.54 $\pm$ 0.28 \\  
    \textbf{RelationNet}     & 87.34 $\pm$ 0.54 & 88.76 $\pm$ 0.65 & 77.65 $\pm$ 0.93 & 77.85 $\pm$ 0.36 & 74.27 $\pm$ 0.54 & 76.93 $\pm$ 0.74 & 82.32 $\pm$ 0.75 & 80.73 $\pm$ 0.29 & 66.21 $\pm$ 0.43 & 67.35 $\pm$ 0.94 \\  
    \textbf{MetaCRL}         & 92.23 $\pm$ 0.47 & 93.42 $\pm$ 0.31 & 82.36 $\pm$ 0.64 & 82.42 $\pm$ 0.57 & 77.24 $\pm$ 0.52 & 78.23 $\pm$ 0.43 & 84.24 $\pm$ 0.54 & 86.36 $\pm$ 0.72 & 69.26 $\pm$ 0.48 & 70.26 $\pm$ 0.84 \\ 
    \textbf{ANIL}         & 89.34 $\pm$ 0.57 & 89.90 $\pm$ 0.44 & 77.81 $\pm$ 0.39 & 79.60 $\pm$ 0.37 & 74.97 $\pm$ 0.03 & 75.47 $\pm$ 0.11 & 79.40 $\pm$ 0.61 & 82.17 $\pm$ 0.27 & 62.49 $\pm$ 0.76 & 64.08 $\pm$ 0.92  \\ 
    \midrule
    \textbf{PATHOSnet-v2} &92.74 $\pm$ 0.64 & 91.12 $\pm$ 0.99  & 80.18 $\pm$ 0.35 & 84.34 $\pm$ 0.47 & 81.94 $\pm$ 0.50 & 80.37 $\pm$ 0.24 & 85.19 $\pm$ 0.98 & 82.46 $\pm$ 0.36 & 69.07 $\pm$ 0.94 & 70.16 $\pm$ 0.49  \\
    \textbf{COGMEN}       & 91.73 $\pm$ 0.93 & 90.31 $\pm$ 0.11 & 79.14 $\pm$ 0.37 & 80.66 $\pm$ 0.58 & 82.01 $\pm$ 0.97 & 80.65 $\pm$ 0.04 & 81.60 $\pm$ 0.28 & 84.24 $\pm$ 0.95 & 67.71 $\pm$ 0.46 & 68.49 $\pm$ 0.17  \\
    \textbf{MARLIN}       & 94.76 $\pm$ 0.14 & 94.34 $\pm$ 0.44 & 80.31 $\pm$ 0.74 & 79.69 $\pm$ 0.48 & 76.19 $\pm$ 0.73 & 78.54 $\pm$ 0.08 & 77.67 $\pm$ 0.81 & 79.96 $\pm$ 0.20 & 70.17 $\pm$ 0.40 & 70.84 $\pm$ 0.73  \\
    \textbf{BMHP} & 94.15 $\pm$ 0.19 & 93.81 $\pm$ 0.34 & 81.09 $\pm$ 0.22 & 80.87 $\pm$ 0.31 & 79.25 $\pm$ 0.24 & 80.25 $\pm$ 0.36 & 85.21 $\pm$ 0.49 & 85.91 $\pm$ 0.53 & 72.39 $\pm$ 0.55 & 73.56 $\pm$ 0.69 \\
    \textbf{FPM-Net} & 95.02 $\pm$ 0.56 & 93.99 $\pm$ 0.27 & 79.23 $\pm$ 0.50 & 78.44 $\pm$ 0.46 & 83.15 $\pm$ 0.56 & 81.09 $\pm$ 0.28 & 84.31 $\pm$ 0.56 & 82.12 $\pm$ 0.65 & 73.09 $\pm$ 0.22 & 72.12 $\pm$ 0.39 \\
    \textbf{TAE} & 95.19 $\pm$ 0.77 & 92.15 $\pm$ 0.45 & 83.51 $\pm$ 0.37 & 89.13 $\pm$ 0.56 & 84.53 $\pm$ 0.39 & 87.20 $\pm$ 0.21 & 86.68 $\pm$ 0.45 & 90.14 $\pm$ 0.68 & 73.81 $\pm$ 0.32 & 74.69 $\pm$ 0.30 \\
    \textbf{CBERL} & 94.30 $\pm$ 0.25 & 93.41 $\pm$ 0.56 & 86.41 $\pm$ 0.65 & 89.88 $\pm$ 0.72 & 86.64 $\pm$ 0.51 & 87.65 $\pm$ 0.55 & 87.84 $\pm$ 0.59 & 87.01 $\pm$ 0.52 & 72.19 $\pm$ 0.89 & 74.64 $\pm$ 0.63 \\
    \textbf{Ours}         & \textbf{96.96 $\pm$ 0.52} & \textbf{96.66 $\pm$ 0.17} & \textbf{87.32 $\pm$ 0.36} & \textbf{93.96 $\pm$ 0.17} & \textbf{86.79 $\pm$ 0.72} & \textbf{92.37 $\pm$ 0.20} & \textbf{89.86 $\pm$ 0.39} & \textbf{91.96 $\pm$ 0.38} & \textbf{73.92 $\pm$ 0.66} & \textbf{76.06 $\pm$ 0.92}  \\    
    \bottomrule
  \end{tabular}}
\end{table*}

\begin{table*}
  \centering
  \caption{Robustness Comparison with Baseline Models on Three Scenarios of CMU-MOSEI, i.e., Fog, Mask, and Distortion. The values in the table represent the Accuracy(\%) with a 95\% confidence interval, and the optimal values will be highlighted in bold.}
  \label{tab:robust}
  \resizebox{\linewidth}{!}{
  \begin{tabular}{l|cccc|cccc|cccc}
    \toprule
    \multirow{2}{*}{\textbf{Methods}} & \multicolumn{4}{c|}{\textbf{Fog}} & \multicolumn{4}{c|}{\textbf{Mask}}  & \multicolumn{4}{c}{\textbf{Distortion}} \\
    & \textbf{0\%} & \textbf{10\%} & \textbf{30\%} & \textbf{50\%} & \textbf{0\%} & \textbf{10\%} & \textbf{30\%} & \textbf{50\%} & \textbf{0\%} & \textbf{10\%} & \textbf{30\%} & \textbf{50\%} \\
    \midrule
    \textbf{FN2EN}        & 84.8 $\pm$ 0.2 & 83.6 $\pm$ 0.3 & 79.5 $\pm$ 0.6 & 76.5 $\pm$ 0.3 & 84.8 $\pm$ 0.2 & 83.5 $\pm$ 0.4 & 80.1 $\pm$ 0.5 & 77.4 $\pm$ 0.3 & 84.8 $\pm$ 0.2 & 82.7 $\pm$ 0.4 & 79.4 $\pm$ 0.5 & 74.6 $\pm$ 0.4 \\
    \textbf{VGG-F-CNN}    & 75.6 $\pm$ 0.7 & 73.8 $\pm$ 0.5 & 70.4 $\pm$ 0.5 & 68.4 $\pm$ 0.3 & 75.6 $\pm$ 0.7 & 73.8 $\pm$ 0.5 & 71.6 $\pm$ 0.7 & 68.7 $\pm$ 0.4 & 75.6 $\pm$ 0.7 & 74.5 $\pm$ 0.6 & 71.5 $\pm$ 0.3 & 66.4 $\pm$ 0.3 \\
    \textbf{PropNet-CNN}  & 72.2 $\pm$ 0.3 & 71.3 $\pm$ 0.4 & 68.8 $\pm$ 0.7 & 65.3 $\pm$ 0.2 & 72.2 $\pm$ 0.3 & 71.6 $\pm$ 0.7 & 69.5 $\pm$ 0.7 & 65.6 $\pm$ 0.7 & 72.2 $\pm$ 0.3 & 71.4 $\pm$ 0.3 & 68.6 $\pm$ 0.4 & 62.5 $\pm$ 0.3 \\
    \textbf{UniMSE}       & 87.0 $\pm$ 0.6 & 86.2 $\pm$ 0.8 & 82.3 $\pm$ 0.6 & 76.4 $\pm$ 0.5 & 87.0 $\pm$ 0.6 & 86.1 $\pm$ 0.5 & 82.3 $\pm$ 0.4 & 77.1 $\pm$ 0.3 & 87.0 $\pm$ 0.6 & 86.0 $\pm$ 0.3 & 82.5 $\pm$ 0.7 & 77.4 $\pm$ 0.6 \\
    \textbf{SepTr}        & 80.4 $\pm$ 0.5 & 78.6 $\pm$ 0.3 & 76.4 $\pm$ 0.5 & 71.4 $\pm$ 0.7 & 80.4 $\pm$ 0.5 & 79.7 $\pm$ 0.2 & 77.0 $\pm$ 0.6 & 73.9 $\pm$ 0.4 & 80.4 $\pm$ 0.5 & 79.3 $\pm$ 0.4 & 76.4 $\pm$ 0.3 & 71.5 $\pm$ 0.6 \\
    \midrule
    \textbf{MAML}         & 81.4 $\pm$ 0.7 & 79.2 $\pm$ 0.4 & 76.7 $\pm$ 0.5 & 73.6 $\pm$ 0.5 & 81.4 $\pm$ 0.7 & 80.8 $\pm$ 0.5 & 77.8 $\pm$ 0.5 & 74.4 $\pm$ 0.2 & 81.4 $\pm$ 0.7 & 79.9 $\pm$ 0.3 & 76.3 $\pm$ 0.2 & 72.4 $\pm$ 0.4 \\
    \textbf{Reptile}      & 83.5 $\pm$ 0.7 & 82.5 $\pm$ 0.7 & 80.4 $\pm$ 0.6 & 77.5 $\pm$ 0.4 & 83.5 $\pm$ 0.7 & 82.4 $\pm$ 0.3 & 80.0 $\pm$ 0.4 & 76.8 $\pm$ 0.6 & 83.5 $\pm$ 0.7 & 82.4 $\pm$ 0.6 & 78.7 $\pm$ 0.7 & 74.3 $\pm$ 0.6 \\
    \textbf{ProtoNet}     & 83.4 $\pm$ 0.5 & 81.4 $\pm$ 0.6 & 80.8 $\pm$ 0.4 & 77.8 $\pm$ 0.3 & 83.4 $\pm$ 0.5 & 82.5 $\pm$ 0.3 & 79.7 $\pm$ 0.4 & 75.4 $\pm$ 0.5 & 83.4 $\pm$ 0.5 & 82.1 $\pm$ 0.4 & 79.4 $\pm$ 0.5 & 73.4 $\pm$ 0.2 \\
    \textbf{RelationNet}  & 82.3 $\pm$ 0.8 & 81.4 $\pm$ 0.5 & 78.1 $\pm$ 0.6 & 74.4 $\pm$ 0.6 & 82.3 $\pm$ 0.8 & 81.4 $\pm$ 0.6 & 79.6 $\pm$ 0.7 & 76.3 $\pm$ 0.4 & 82.3 $\pm$ 0.8 & 81.2 $\pm$ 0.3 & 77.4 $\pm$ 0.8 & 83.6 $\pm$ 0.4 \\ 
    \textbf{MetaCRL}      & 84.2 $\pm$ 0.5 & 83.4 $\pm$ 0.7 & 81.7 $\pm$ 0.5 & 78.6 $\pm$ 0.7 & 84.2 $\pm$ 0.5 & 83.5 $\pm$ 0.4 & 82.4 $\pm$ 0.6 & 79.4 $\pm$ 0.3 & 84.2 $\pm$ 0.5 & 83.1 $\pm$ 0.7 & 80.2 $\pm$ 0.4 & 76.7 $\pm$ 0.3 \\
    \textbf{ANIL}         & 79.4 $\pm$ 0.6 & 78.3 $\pm$ 0.4 & 75.5 $\pm$ 0.4 & 72.5 $\pm$ 0.7 & 79.4 $\pm$ 0.6 & 78.1 $\pm$ 0.5 & 75.9 $\pm$ 0.4 & 72.6 $\pm$ 0.3 & 79.4 $\pm$ 0.6 & 78.0 $\pm$ 0.4 & 75.1 $\pm$ 0.3 & 70.4 $\pm$ 0.2 \\
     \midrule
    \textbf{PATHOSnet-v2} & 85.2 $\pm$ 1.0 & 84.1 $\pm$ 0.2 & 81.8 $\pm$ 0.3 & 76.4 $\pm$ 0.4 & 85.2 $\pm$ 1.0 & 83.7 $\pm$ 0.7 & 80.5 $\pm$ 0.4 & 75.4 $\pm$ 0.3 & 85.2 $\pm$ 1.0 & 84.6 $\pm$ 0.5 & 81.5 $\pm$ 0.6 & 76.2 $\pm$ 0.4 \\
    \textbf{COGMEN}       & 81.6 $\pm$ 0.3 & 79.6 $\pm$ 0.4 & 75.6 $\pm$ 0.3 & 71.3 $\pm$ 0.2 & 81.6 $\pm$ 0.3 & 79.8 $\pm$ 0.4 & 76.4 $\pm$ 0.6 & 72.5 $\pm$ 0.7 & 81.6 $\pm$ 0.3 & 80.2 $\pm$ 0.4 & 77.6 $\pm$ 0.3 & 71.7 $\pm$ 0.8 \\
    \textbf{MARLIN}       & 77.7 $\pm$ 0.8 & 76.3 $\pm$ 0.7 & 73.2 $\pm$ 0.4 & 71.4 $\pm$ 0.3 & 77.7 $\pm$ 0.8 & 76.5 $\pm$ 0.6 & 73.6 $\pm$ 0.4 & 70.6 $\pm$ 0.4 & 77.7 $\pm$ 0.8 & 76.7 $\pm$ 0.4 & 73.4 $\pm$ 0.3 & 67.3 $\pm$ 0.5 \\
    \textbf{BMHP} & 85.2 $\pm$ 0.5 & 84.0 $\pm$ 0.4 & 82.3 $\pm$ 0.4 & 79.2 $\pm$ 0.5 & 85.2 $\pm$ 0.5 & 83.9 $\pm$ 0.5 & 81.6 $\pm$ 0.4 & 75.2 $\pm$ 0.4 & 85.2 $\pm$ 0.5 & 84.0 $\pm$ 0.6 & 82.1 $\pm$ 0.7 & 79.5 $\pm$ 0.5 \\
    \textbf{FPM-Net} & 84.3 $\pm$ 0.6 & 83.2 $\pm$ 0.5 & 81.9 $\pm$ 0.5 & 78.6 $\pm$ 0.4 & 84.3 $\pm$ 0.6 & 82.8 $\pm$ 0.5 & 79.7 $\pm$ 0.6 & 75.0 $\pm$ 0.5 & 84.3 $\pm$ 0.6 & 83.1 $\pm$ 0.6 & 80.4 $\pm$ 0.5 & 74.9 $\pm$ 0.6 \\
    \textbf{TAE} & 86.6 $\pm$ 0.5 & 85.0 $\pm$ 0.5 & 83.7 $\pm$ 0.4 & 81.6 $\pm$ 0.7 & 84.3 $\pm$ 0.6 & 83.2 $\pm$ 0.3 & 81.9 $\pm$ 0.7 & 78.0 $\pm$ 0.6 & 84.3 $\pm$ 0.6 & 83.5 $\pm$ 0.6 & 81.9 $\pm$ 0.5 & 78.0 $\pm$ 0.2 \\
    \textbf{CBERL} & 87.8 $\pm$ 0.6 & 87.2 $\pm$ 0.4 & 86.7 $\pm$ 0.6 & 84.3 $\pm$ 0.4 & 84.3 $\pm$ 0.6 & 82.7 $\pm$ 0.9 & 80.1 $\pm$ 0.2 & 76.1 $\pm$ 0.5 & 84.3 $\pm$ 0.6 & 83.5 $\pm$ 0.7 & 81.0 $\pm$ 0.7 & 78.2 $\pm$ 0.4 \\
    \textbf{Ours}         & \textbf{89.8 $\pm$ 0.4} & \textbf{89.3 $\pm$ 0.3} & \textbf{87.9 $\pm$ 0.4} & \textbf{85.1 $\pm$ 0.4} & \textbf{89.8 $\pm$ 0.4} & \textbf{87.3 $\pm$ 0.4} & \textbf{84.6 $\pm$ 0.2} & \textbf{80.4 $\pm$ 0.4} & \textbf{89.8 $\pm$ 0.4} & \textbf{88.1 $\pm$ 0.4} & \textbf{86.3 $\pm$ 0.4} & \textbf{83.4 $\pm$ 0.3} \\
    \bottomrule
  \end{tabular}}
\end{table*}

\begin{figure*}
    \begin{minipage}{0.48\textwidth}
    \centering
    \includegraphics[width=\textwidth]{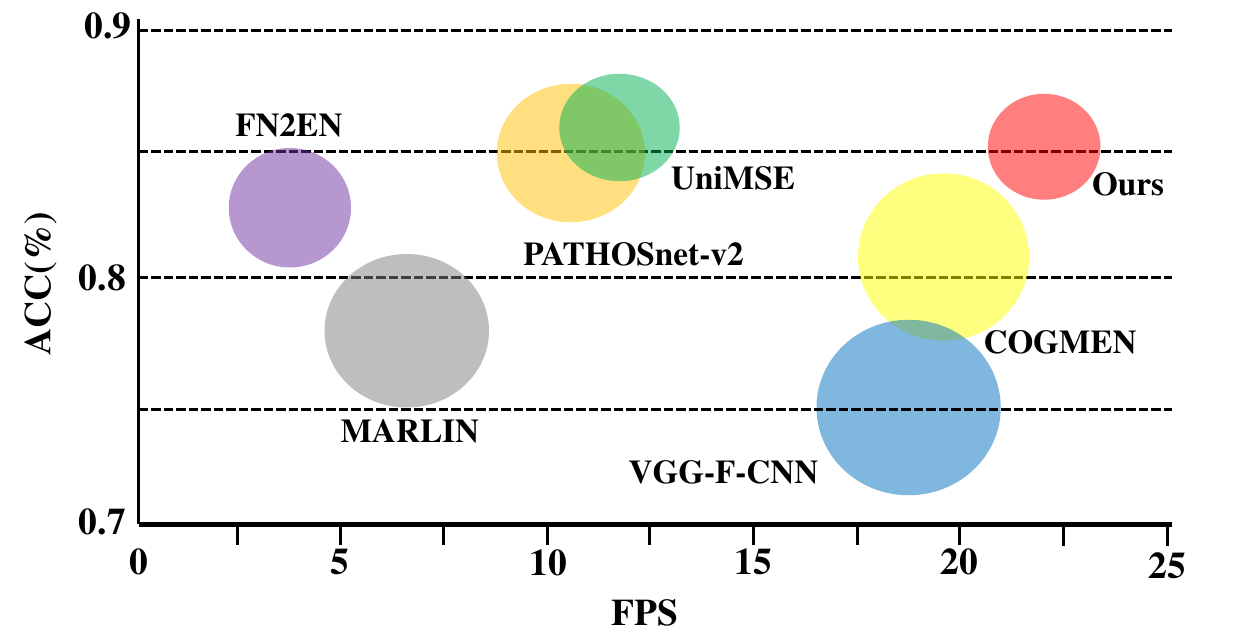}
    \caption{Trade-off performance. It provides the comparative results of models’ ability to balance performance and efficiency.}
    \label{fig:ex1}
    \end{minipage}
    \hfill
    \begin{minipage}{0.48\textwidth}
    \centering
    \includegraphics[width=\textwidth]{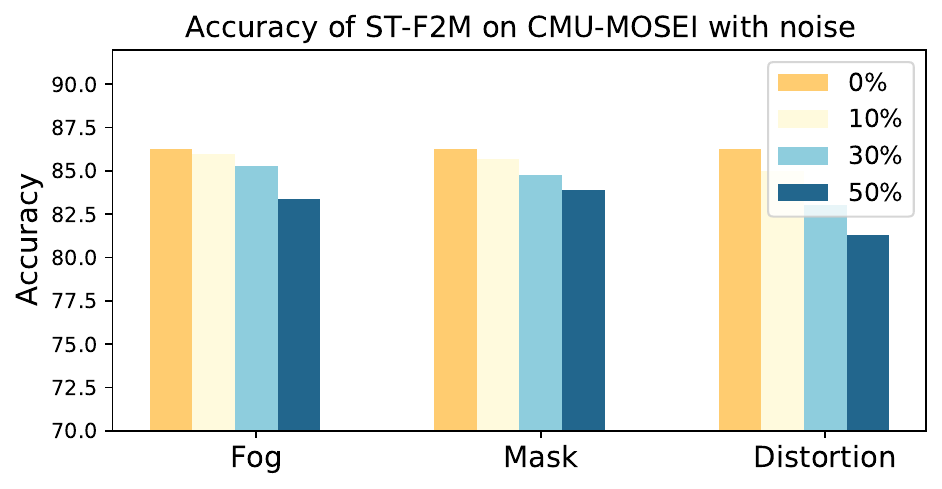}
    \caption{Robust analysis on CMU-MOSEI. (\%) represents the proportion of manually introduced noise in the dataset.}
    \label{fig:ex2}
    \end{minipage}
\end{figure*}

\begin{figure*}
    \begin{minipage}{0.38\textwidth}
        \centering
        \includegraphics[width=\linewidth]{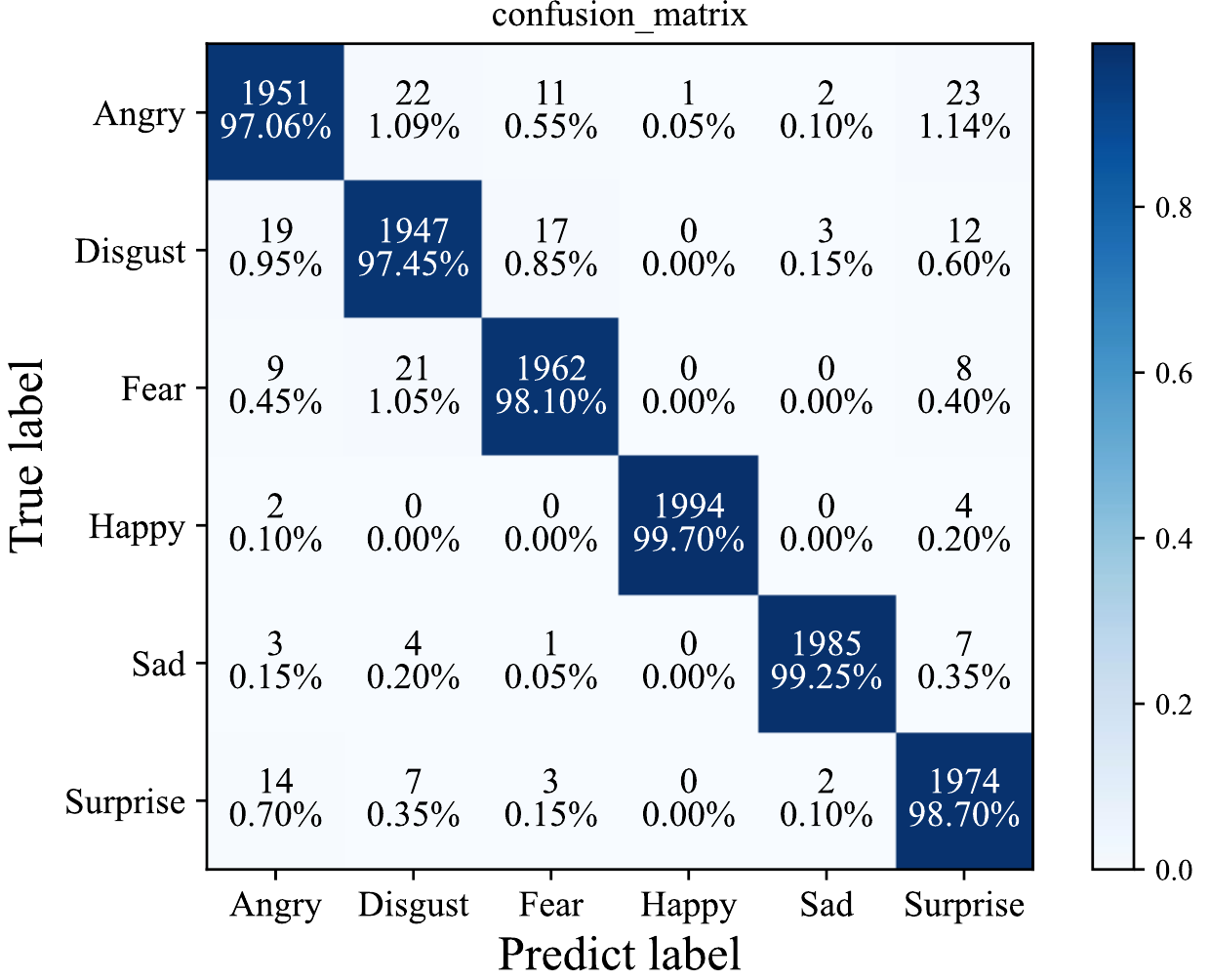}
        \caption{The confusion matrix of six emotions.}
        \label{fig:ex3}
    \end{minipage}
    \hfill
    \begin{minipage}{0.59\textwidth}
        \centering
        \includegraphics[width=\linewidth]{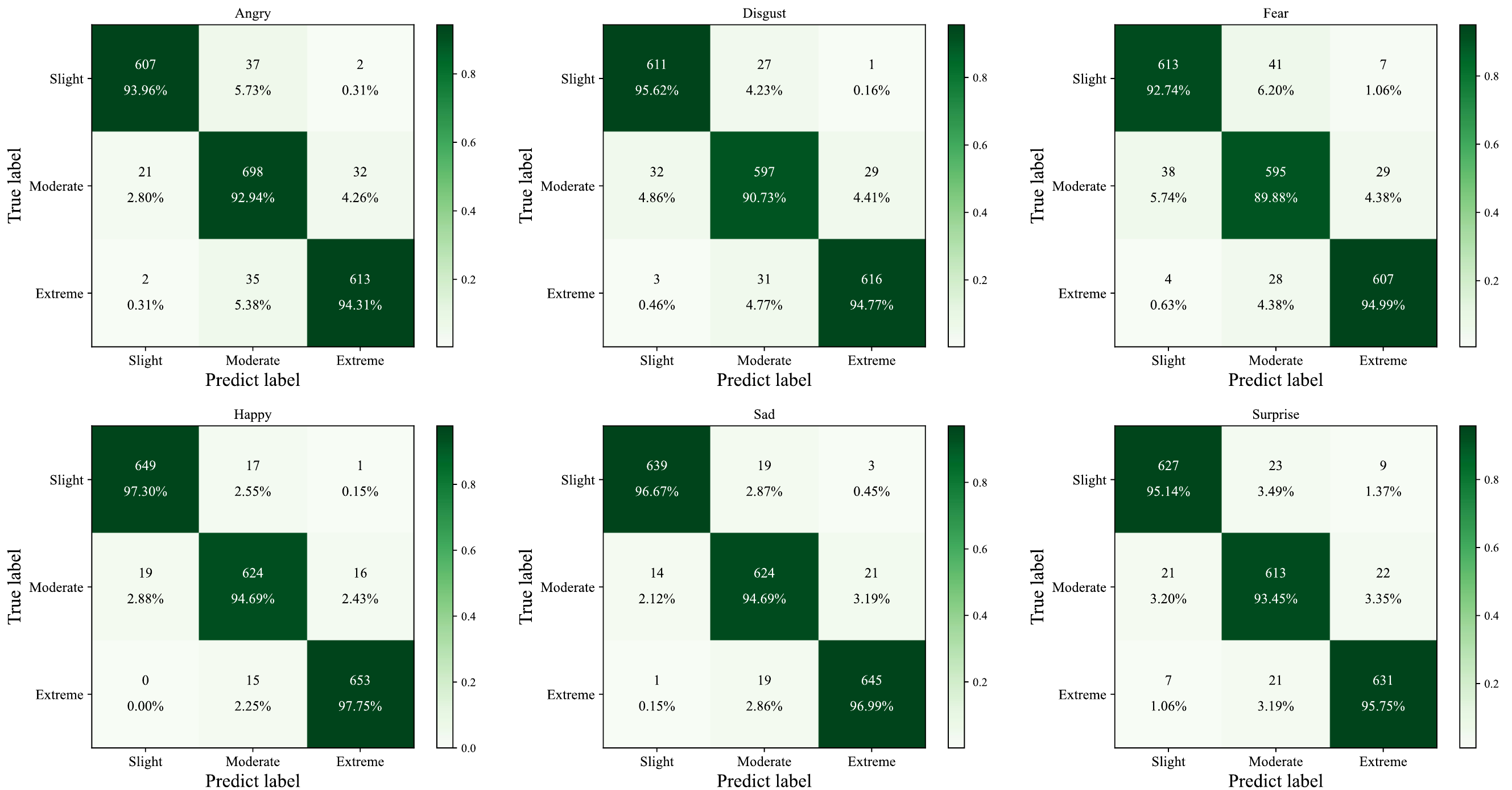}
        \caption{The confusion matrices of the intensities of six emotions.}
        \label{fig:ex4}
    \end{minipage}
\end{figure*}

\begin{table}
  \centering
  \caption{Performance comparison when facing mislabeling with few-shot settings on CK+. ``Standard'' indicates the general settings as in Subsection \ref{sec:5.3}, and ``Few-Shot'' indicates the few-shot settings with 30\% mislabeled samples.}
  \label{tab:robust_few-shot}
  \resizebox{\linewidth}{!}{
  \begin{tabular}{l|cc|cc}
    \toprule
    \multirow{2}{*}{\textbf{Methods}} & \multicolumn{2}{c|}{\textbf{Standard}} & \multicolumn{2}{c}{\textbf{Few-Shot}} \\
    & \textbf{Average ACC} & \textbf{Worst ACC} & \textbf{Average ACC} & \textbf{Worst ACC} \\
    \midrule
    \textbf{MAML}         & 85.70 $\pm$ 0.64 & 83.16 $\pm$ 0.59 & 49.14 $\pm$ 0.56 & 45.12 $\pm$ 0.77 \\
    \textbf{ProtoNet}         & 86.40 $\pm$ 0.67 & 84.21 $\pm$ 0.58 & 48.15 $\pm$ 0.62 & 44.37 $\pm$ 0.60 \\
    \textbf{PATHOSnet-v2} & 92.74 $\pm$ 0.64 & 91.30 $\pm$ 0.44 & 43.15 $\pm$ 0.27 & 40.50 $\pm$ 0.36 \\
    \textbf{MARLIN}       & 94.76 $\pm$ 0.14 & 92.8 $\pm$ 0.36 & 45.41 $\pm$ 0.22 & 42.01 $\pm$ 0.63 \\
    \textbf{BMHP} & 94.15 $\pm$ 0.19 & 90.69 $\pm$ 0.25 & 48.58 $\pm$ 0.32 & 42.79 $\pm$ 0.45 \\
    \textbf{FPM-Net} & 95.02 $\pm$ 0.56 & 93.46 $\pm$ 0.54 & 49.28 $\pm$ 0.32 & 47.28 $\pm$ 0.38 \\
    \textbf{TAE} & 95.19 $\pm$ 0.77 & 91.49 $\pm$ 0.84 & 50.41 $\pm$ 0.65 & 47.72 $\pm$ 0.58 \\
    \textbf{CBERL} & 94.30 $\pm$ 0.25 & 91.76 $\pm$ 0.18 & 51.41 $\pm$ 0.83 & 48.22 $\pm$ 0.78 \\
    \textbf{Ours}         & \textbf{96.96 $\pm$ 0.52} & \textbf{95.01 $\pm$ 0.39} & \textbf{57.46 $\pm$ 0.50} & \textbf{55.80 $\pm$ 0.47} \\   
    \bottomrule
  \end{tabular}}
\end{table}

\begin{figure*}
    \begin{minipage}{0.48\textwidth}
        \centering
        \includegraphics[width=\textwidth]{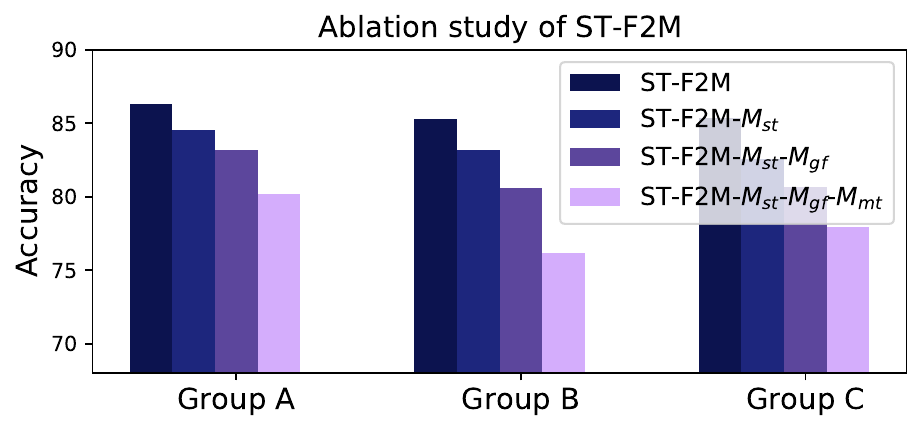}
        \caption{Ablation study of ST-F2M.}
        \label{fig:ex5}
    \end{minipage}
    \hfill
    \begin{minipage}{0.48\textwidth}
        \centering
        \includegraphics[width=\textwidth]{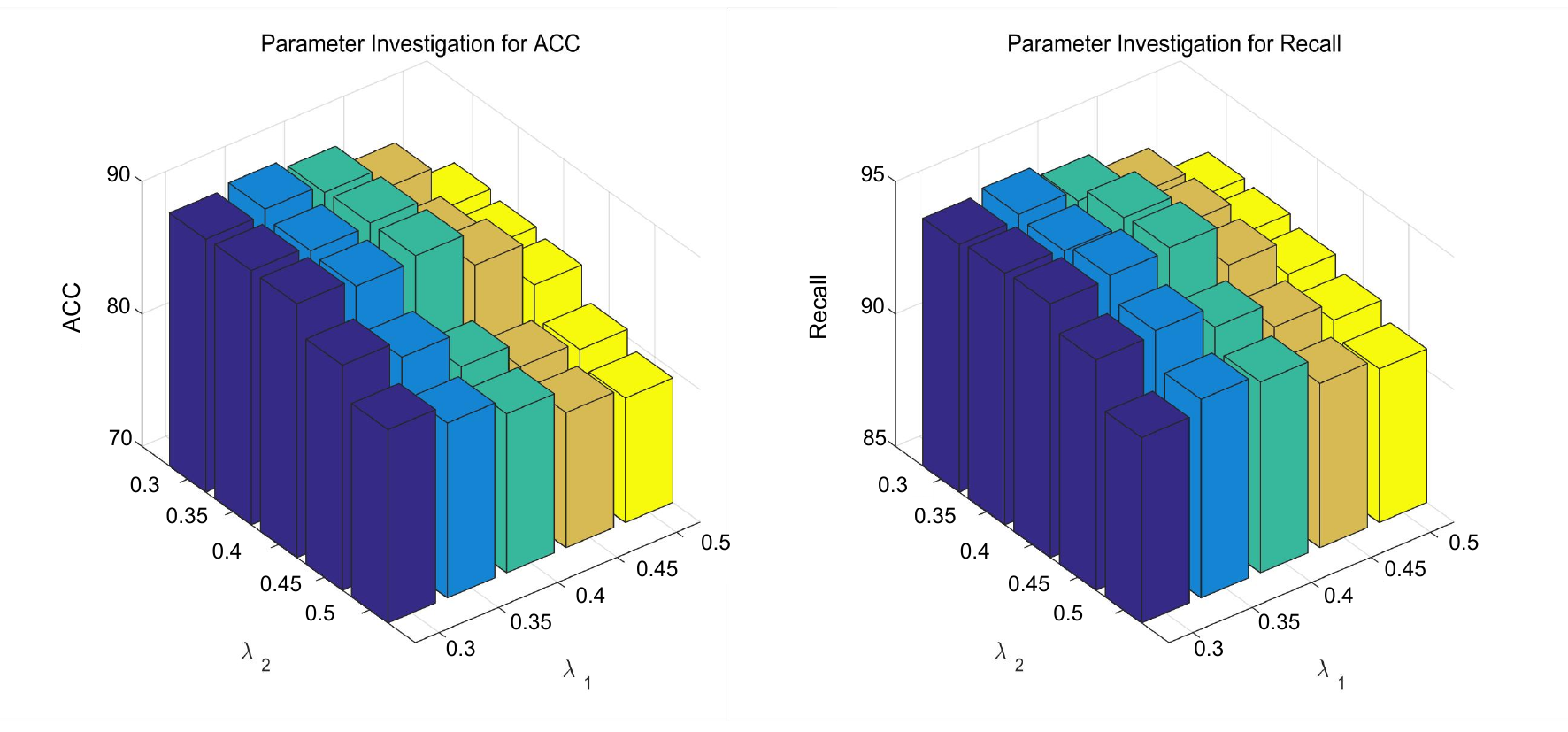}
        \caption{Parameter investigations about $\left \{ \lambda _{1},\lambda _{2} \right \} $.}
        \label{fig:ex6}
    \end{minipage}
\end{figure*}

\begin{figure*}
    \centering
    \subfigure[Data collection]{\includegraphics[width=0.34\linewidth]{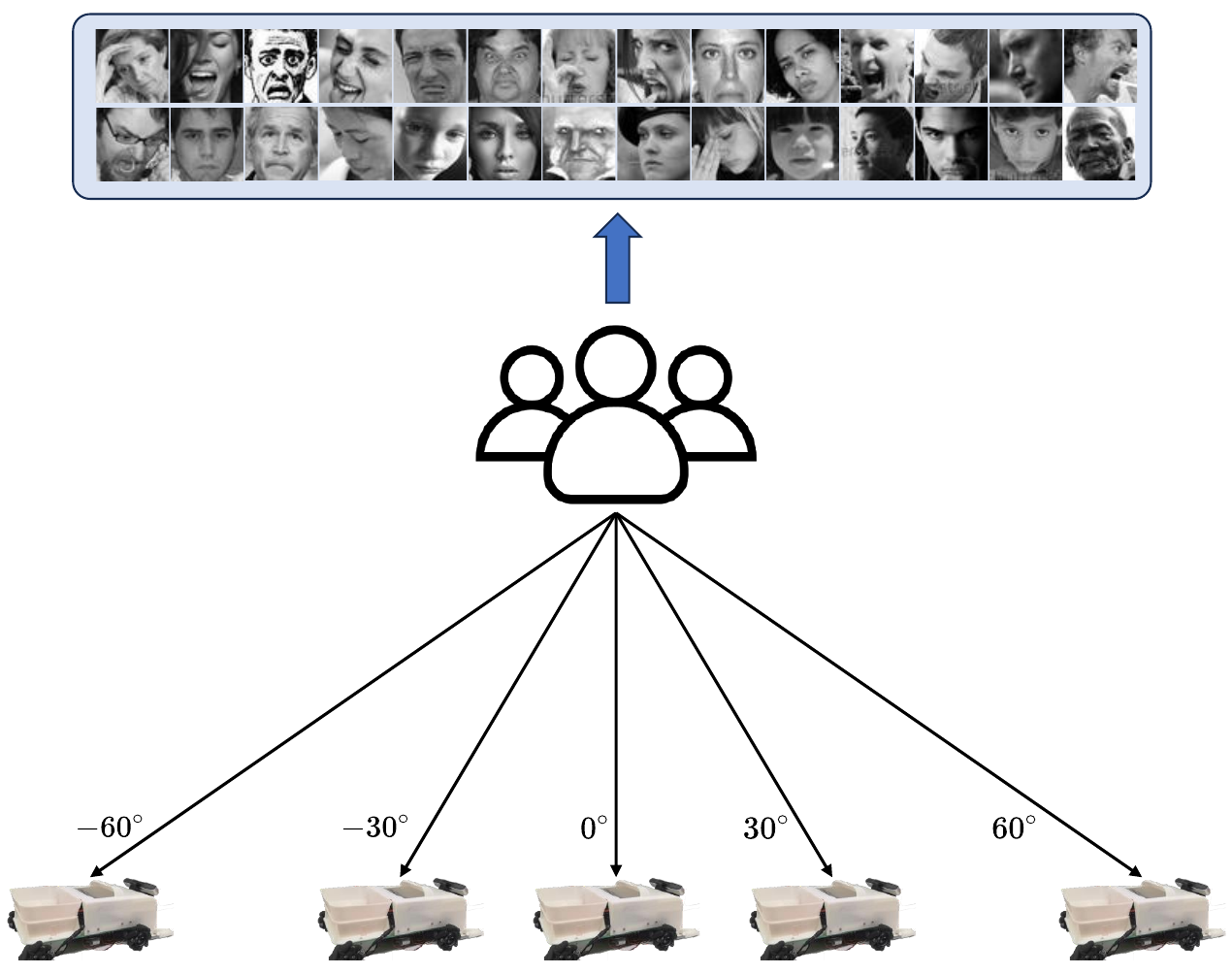}\label{fig:robot-result_1}}
    \hfill
     \subfigure[Comparison results]{\includegraphics[width=0.63\linewidth]{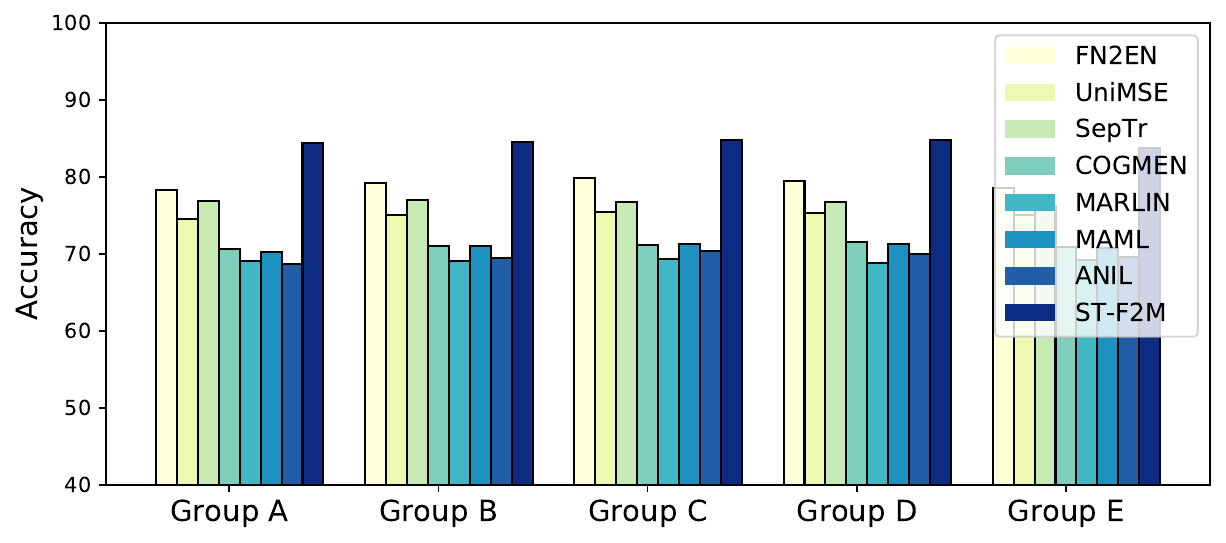}\label{fig:robot-result_2}}
    \caption{Comparison results of practical case study. Group A $\sim $ Group E respectively represent the five groups of data collected in the robot platform. For each set of data, the robot shoots and identifies from different angles, that is, the same direction as the tester faces $-60^\circ $, $-30^\circ $, $0^\circ $, $30^\circ $, and $60^\circ $. (a) shows the data collection environment, and (b) provides the comparison results.}
    \label{fig:robot-result}
\end{figure*}

\section{Experiments}
\label{sec:5}
In this section, we first introduce the experimental setup in Subsection \ref{sec:5.1}. Then, we present the results of performance comparison, computational efficiency comparison, and robustness analysis in Subsections \ref{sec:5.2} to \ref{sec:5.4}, demonstrating the superiority of ST-F2M. Next, we conduct ablation studies and parameter investigation to further explore how ST-F2M works and what makes ST-F2M work well in Subsections \ref{sec:5.5} to \ref{sec:5.6} respectively. Finally, we evaluate the practical application effect of ST-F2M in Subsection \ref{sec:5.7}.

\subsection{Experimental Setup}
\label{sec:5.1}
In this subsection, we introduce implementation details, datasets, and baselines of our experiments in turn.

\subsubsection{\textbf{Implementation Details}}
\label{sec:5.1.1}
We opt for the most commonly used ResNet-18 as the backbone. After the convolution and filtering stages, we sequentially apply batch normalization, ReLU non-linear activation, and $2 \times 2$ max pooling (achieved through stride convolutions). The last layer ultimately produces two separate channels as its final output. The first channel is fed into the proposed fuzzy systems, i.e., FCIS and FKIS, to extract fuzzy semantic information, while the second channel is directed towards a softmax layer. These network architectures undergo an initial pre-training phase and remain unaltered throughout the training process. As we transition to the optimization stage, we utilize the Adam optimizer \cite{tato2018improving} for training our model. We set the values of momentum and weight decay to 0.9 and $10^{-4}$, respectively. The initial learning rate for all experiments is set at 0.2, with the flexibility for linear scaling as needed. All experiments are executed on 10 NVIDIA V100 GPUs.

\subsubsection{\textbf{Datasets}}
\label{sec:5.1.2}
We evaluate our ST-F2M on five benchmark datasets, including: (i) CK+ \cite{lucey2010extended}, which contains 593 video sequences from 123 different subjects. The videos are recorded at 30 frames per second (FPS) with $640 \times  490$ or $640 \times  480$ pixel resolution; (ii) DISFA \cite{mavadati2013disfa} contains 27 videos of 4844 frames, which is mainly used for facial expression-based FER; (iii) WFLW \cite{wu2018look}, which contains 10,000 samples and the sentiment intensity is added by manual annotation. The rich attribute annotations of this dataset, such as occlusion and blur, are significantly helpful in simulating defects in real-life; (iv) CMU-MOSEI \cite{zadeh2018multimodal}, which contains 22,856 video clips from 1000 different speakers; and (v) CREMA-D \cite{cao2014crema} consists of 7,442 video clips that employ both facial and vocal data to categorize six fundamental emotional states. Following \cite{zhang2022tailor}, we pre-extract 35-dimensional visual features from video frames through FACET \cite{baltruvsaitis2016openface}, and pre-extract 300-dimensional text features from video text through Glove \cite{pennington2014glove} to supplement the missing modal information. Note that for the facial expression datasets, DISFA and WFLW, we assign emotion and intensity information to the training data based on generalized fuzzy rules, and named the datasets DISFA-FER and WFLW-FER which will be open sourced.

\subsubsection{\textbf{Baselines}}
\label{sec:5.1.3}
We introduce four types of baselines: (i) state-of-the-art methods (before 2024) of each dataset, including FN2EN \cite{ding2017facenet2expnet}, VGG-F \cite{bulat2022pre}, PropNet-CNN \cite{huang2020propagationnet}, UniMSE \cite{hu2022unimse}, and SepTr \cite{ristea2022septr}; (ii) meta-learning methods, including MAML \cite{finn2017model}, Reptile \cite{reptile}, ProtoNet \cite{protonet}, RelationNet \cite{relationnet}, MetaCRL \cite{metacrl}, and ANIL \cite{raghu2019rapid}; and (iii) prominent and recently proposed multi-modal FER methods, including 
PATHOSnet-v2 \cite{scotti2021combining}, COGMEN \cite{joshi2022cogmen}, and MARLIN \cite{cai2023marlin}, BMHP \cite{zhu2024emotion}, FPM-Net \cite{fan2024fusing}, TAE \cite{cheng2024novel}, and CBERL \cite{meng2024deep}. Note that for single-modal methods under multi-modal data sets, we use the bagging \cite{oza2001online} method to integrate after solving the single modality. For the meta-learning method, we adopt the same task construction method as ST-F2M for multi-modal emotion recognition.

\subsection{Performance Comparison}
\label{sec:5.3}
To evaluate the effectiveness of ST-F2M, we conduct comparative experiments with the baselines on the five datasets mentioned in Section \ref{sec:5.1}, containing six basic emotions, i.e., happiness, anger, sadness, disgust, fear, and surprise, and each emotion consists of three intensity levels, i.e., low, medium and high. In this experiment, we set two metrics, including accuracy (ACC) and recall (RECALL). We record the average value of 10 rounds to obtain the results of these two metrics. 

The results are shown in Table \ref{tab:ex1}. From the results, we can observe that across all datasets, our approach consistently outperforms both meta-learning methods and multi-modal facial expression recognition methods. Moreover, our ST-F2M achieves results similar to state-of-the-art (SOTA) baselines which rely on large datasets, while our ST-F2M only has limited data support. This indicates that ST-F2M can achieve similar or even superior generalization improvements compared to baselines, without the need for extensive data reliance. It is worth noting that our method is always the SOTA method on the Recall, which illustrates its superiority in fields such as disease diagnosis that require high recall rates \cite{battineni2020applications}.

\subsection{Computational Efficiency Comparison}
\label{sec:5.4}
To analyze the computational efficiency of ST-F2M, we construct comparative experiments between ST-F2M and baselines to evaluate their trade-off between performance and efficiency. Specifically, we built an edge computing platform, where the host is NVIDIA Jetson AGX Xavier, including a 512-core NVIDIA Volta GPU with 64 Tensor cores, two NVIDIA deep learning gas pedals, two vision gas pedals, and an eight-core NVIDIA Carmel Arm CPU. We record the running time (FPS) of each model under different accuracy effects.

Figure \ref{fig:ex1} illustrates the results of the trade-off between performance and efficiency. The center of each circle in the figure represents the average result for the respective models, and the circle's area represents a 90\% confidence interval. It's important to note that our method doesn't require additional preprocessing steps and achieves a real-time speed of 20.3 FPS (0.02 seconds per frame). On the same platform, ST-F2M achieves nearly the fastest detection while maintaining a high level of accuracy. This result highlights the superiority of ST-F2M in terms of computational efficiency.

\subsection{Robustness Analysis}
\label{sec:5.2}

\subsubsection{\textbf{When Facing Distortion and Noise}}
Emotion recognition in real life can encounter various interferences. Take medical environments as an example, the emotional data of the patients may have large portions obscured due to wearing masks or bandages. A robust method should still achieve reliable emotion recognition in the presence of various interferences. To assess the robustness of ST-F2M, we manually introduce multiple noises into the data to simulate real-world interferences, including adding fog \cite{kim2018effective}, masks \cite{ghanbari2020new}, and distortion \cite{luo2020distortion} for visual modality and adding masks for textual modality. All types of noise are introduced into the data with the levels of 10\%, 30\%, and 50\%, respectively. 
Specifically, in the case of fogging, these levels also correspond to increasing fog density; in the case of masking, these levels indicate the proportion of the sample area that is masked; in the case of distortion, these levels refer to the proportion of samples that affected. 

The results are shown in Figure \ref{fig:ex2}. We record the results of ST-F2M on CMU-MOSEI. The accuracy fluctuates within 4\%, demonstrating excellent stability. Table \ref{tab:robust} shows the comparison results of robustness analysis. The results show that ST-F2M obtains relatively stable results under all data sets and scenarios, while other methods suffer from data noises. Therefore, ST-F2M shows superior robustness on real-life FER.

To gain a more intuitive understanding of ST-F2M's performance, we randomly select 2000 cases for each emotion from video data containing 20\% masks. Each case comprises an image and a text message. Figure \ref{fig:ex3} displays the confusion matrix for the six basic emotions (without intensity), while Figure \ref{fig:ex4} presents the confusion matrix for the six emotions with intensity. It is evident that the results on the diagonal show a significant advantage, indicating the robustness of ST-F2M.

\subsubsection{\textbf{When Facing Mislabeling with Few Samples}}
To further assess the robustness of the proposed method, we conduct an experiment targeting scenarios with limited samples and mislabeled data. Using CK+ as the benchmark dataset, we follow the same experimental setup as in Subsection \ref{sec:5.3}, except that each training batch contains only 20\% of the original data to simulate a few-shot setting. Additionally, 30\% of the sampled training data have their emotion labels randomly shuffled to mimic real-world weak supervision with label noise. We report both the average and worst-case test performance of different models in Table \ref{tab:robust_few-shot}. The results show that our method consistently achieves superior performance under these challenging conditions, demonstrating its effectiveness. This also confirms that the fuzzy system accurately models fundamental emotional components and effectively mitigates the impact of mislabeled data during training.

\subsection{Ablation Study}
\label{sec:5.5}
We conduct an ablation study to explore the impact of different modules in ST-F2M, i.e., the spatio-temporal convolution module $M_{st}$, the generalized fuzzy inference module $M_{gf}$, and the meta-optimization module $M_{mt}$. We evaluate their influence on the ST-F2M performance by replacing or redesigning these modules. For instance, $M_{st}$ is replaced with a Conv4-based encoder without considering spatio-temporal features, $M_{gf}$ is disabled, and $M_{mt}$ is substituted with single-level training from scratch. We conduct experiments in three different environments, including CMU-MOSEI (Group A), CMU-MOSEI with 30\% interference (Group B), and CMU-MOSEI with randomly altered speeds (Group C). The latter two are used to assess the impact of ST-F2M on spatial and temporal heterogeneity. 

The results are shown in Figure \ref{fig:ex5}. The results indicate that all three modules significantly enhance the model's performance. Furthermore, they bring noticeable performance improvements, especially in Groups B and C, with $M_{st}$ being particularly prominent. Therefore, our design demonstrates foresight, while highlighting our concern for the issue of spatial and temporal heterogeneity in fine-grained emotion recognition.

\subsection{Parameter Investigation}
\label{sec:5.6}

The impact of hyperparameters on performance cannot be overlooked. In the case of ST-F2M, the description of membership functions in FCIS and FKIS directly affects emotion recognition results. Therefore, we conduct experiments specifically focusing on the attribute membership range $\lambda_1$ in FCIS and the membership range $\lambda_2$ in FKIS to find the optimal settings. The example results are presented in a 3D histogram in Figure \ref{fig:ex6}. Example results are presented as a 3D histogram in figure \ref{fig:ex6}. Specifically, we take the high-intensity value sample of angry emotion as an example to calculate the membership range of the generalized fuzzy inference module when the accuracy and recall rates are the highest. The best effect is when $\lambda_1$ and $\lambda_2$ are both 0.4, which is also the setting of our proposed fuzzy systems. The results indicate that the membership functions we constructed for the system in Subsection \ref{sec:3.2} deliver the best performance. This demonstrates the rigor of our work.

\subsection{Practical Case Study}
\label{sec:5.7}

Although current FER methods achieve good results on benchmark datasets, they have not been deployed and tested in real-life applications \cite{chen2021multimodal, kabir2021emocov, zhang2021customer}. There is a big difference between the real scenarios and the benchmark datasets, and it is interfered with by various aspects \cite{wang2020learning, poria2017review}. Although the aforementioned experiments have proven the outstanding effect and robustness of ST-F2M in different scenarios, to evaluate the effect of the model in actual deployment, we further select more complex medical robot scenarios for experiments.

We use the same implementation details as Subsection \ref{sec:5.1} but a different deployment platform, i.e., a self-built robot platform \footnote{More details on the robot build, please see: \url{https://github.com/WangJingyao07/PickBallBot-5mao/tree/main}}. The robot host we use is NVIDIA Jetson AGX Xavier, including a 512-core NVIDIA Volta GPU with 64 Tensor cores, two NVIDIA deep learning accelerator pedals, two visual accelerator pedals, and an eight-core NVIDIA Carmel Arm CPU. Our ST-F2M is experimented on a unified model edge computing platform and ultimately embedded on a medical robotics platform. We set up the communication and underlying control algorithms to achieve a complete system-closed loop. In this experiment, we use the built robot entity to conduct 5 groups of experiments in the laboratory simulation environment. The data of each group contains 20 pieces of data with different angles. Figure \ref{fig:robot-result_1} shows the data collection. Each piece of data is a 30-second video, labeled with the emotional sequence contained in the video. The robot's goal is to recognize the full range of emotions contained in video data. Note that the data in this scene may have various interferences, such as camera distortion and object occlusion \cite{tang2017precision, dong2016occlusion}. Figure \ref{fig:intro_3} provides examples of real-world FER samples. 

From Figure \ref{fig:robot-result}, we can observe that: (i) ST-F2M still achieves superior results in real-life deployment; (ii) other FER methods have unsatisfactory results in the robot entity test environment, possibly due to various reasons like camera distortion, object occlusion, etc.; and (iii) ST-F2M can still obtain relatively stable FER results when facing many problems caused by the robot entity testing environment. These observations prove the advantages of ST-F2M in practical applications and illustrate the effectiveness and forward-lookingness of our design for the real-life FER field.

\section{Conclusion}
\label{sec:6}

In this paper, we propose a novel Spatio-Temporal Fuzzy-oriented Multi-modal Meta-learning (ST-F2M) framework for real-life fine-grained emotion recognition in videos. It addresses three key challenges in this field. Specifically, considering spatiotemporal heterogeneity, we integrate temporal and spatial convolutions to encode multi-modal emotional states in long videos. Then, we design i) a fuzzy emotion and fuzzy knowledge inference system based on generalized fuzzy rules; and ii) a fuzzy-oriented multi-modal meta-learning paradigm to achieve fast and robust fine-grained emotion recognition. Comprehensive experiments on five FER datasets demonstrate the robustness and generalization of ST-F2M. In practical applications, our method has been successfully deployed on edge computing platforms such as medical robots, achieving robust fine-grained emotion recognition. In addition, we construct two FER datasets, namely DISFA-FER and WFLW-FER, based on the constructed generalized fuzzy inference system. We hope that ST-F2M can provide valuable insights into other application areas of FER.




%




\ifCLASSOPTIONcaptionsoff
  \newpage
\fi



%


\bibliographystyle{IEEEtran}
\bibliography{reference}

\end{document}